\documentclass{article}

\usepackage[preprint]{neurips_2026}
\usepackage{array}
\usepackage{colortbl}
\usepackage{xurl}
\usepackage{color}
\usepackage{soul}
\usepackage{siunitx}
\usepackage{listings}
\usepackage{subcaption}
\usepackage{float}
\usepackage[normalem]{ulem}
\usepackage{setspace}
\usepackage{placeins}
\usepackage{afterpage}
\usepackage{hhline}
\usepackage{enumitem}
\usepackage{pifont}
\usepackage{arydshln}
\usepackage{adjustbox}
\usepackage{lipsum}
\usepackage{tabularx}
\usepackage{makecell}
\usepackage{wrapfig}
\usepackage{moreverb}
\usepackage{multirow}
\usepackage{xspace}
\usepackage{hyperref}
\usepackage{epsfig}
\usepackage{endnotes}
\usepackage{url}
\usepackage{fancyhdr}
\usepackage{xcolor}
\usepackage{caption}
\usepackage{amsfonts}
\usepackage{amsmath}
\usepackage[framemethod=TikZ]{mdframed}
\definecolor{codegray}{RGB}{128, 128, 128}
\definecolor{backcolor}{RGB}{250, 250, 250}
\definecolor{rulecolor}{RGB}{120, 120, 120}
\definecolor{keywordcolor}{RGB}{218, 75, 88}
\definecolor{stringcolor}{RGB}{64, 135, 39}
\definecolor{identifiercolor}{RGB}{50, 50, 50}

\usepackage{booktabs}

\lstdefinestyle{mystyle} {
    commentstyle=\color{rulecolor},
    keywordstyle=\color{keywordcolor},
    stringstyle=\color{stringcolor},
    basicstyle=\ttfamily\scriptsize,
    identifierstyle=\color{identifiercolor},
    backgroundcolor=\color{backcolor},
    breakatwhitespace=false,
    breakindent=0pt,
    breaklines=true,
    captionpos=b,
    keepspaces=true,
    numbers=left,
    numberstyle=\ttfamily\tiny\color{rulecolor},
    numbersep=3pt,
    showspaces=false,
    showstringspaces=false,
    showtabs=false,
    framexleftmargin=0pt,
    frame=lines,
    rulecolor=\color{rulecolor},
    rulesepcolor=\color{gray},
    xleftmargin=0pt,
    xrightmargin=0pt,
}
\definecolor{blueframecolor}{RGB}{173,212,251} 
\definecolor{textgray}{gray}{0.5}
\definecolor{typeflyblue}{RGB}{95,148,247} 
\definecolor{typeflyred}{RGB}{202,85,92}
\definecolor{typeflygreen}{RGB}{82,133,54}
\definecolor{typeflypurple}{RGB}{189,21,240}

\newmdenv[  
  linecolor=blueframecolor,
  outerlinewidth=0.5pt, 
  roundcorner=3pt, 
  backgroundcolor=blueframecolor!10,
  frametitlerule=true,
  innertopmargin=3pt,
  innerbottommargin=3pt,
  font=\small
]{customframe}

\definecolor{lightgray}{gray}{0.9}
\definecolor{lightblue}{rgb}{0.9,0.9,1}
\definecolor{blue_bg}{rgb}{0.85,0.85,1}
\definecolor{lightyellow}{rgb}{1,1,0.8}
\definecolor{lightpurple}{rgb}{1,0.85,1}
\definecolor{red}{rgb}{1,0,0}
\definecolor{darkgreen}{rgb}{0.4,0.7,0.3}
\definecolor{lightcyan}{rgb}{0.4, 0.8, 0.9}

\usepackage{tcolorbox}

\newcommand{\remove}[1]{}

\newcommand\sysname{\textit{VeXact}\xspace}

\providecommand{\Description}[1]{}

\title{Diagnosing Training-Inference Mismatch in LLM Reinforcement Learning via a Zero-Mismatch Reference}

\begin{document}

\author{%
  \normalfont
  \textbf{Tianle Zhong$^{1,2,*}$ \quad
  Neiwen Ling$^{1,*}$ \quad
  Yifan Pi$^1$ \quad
  Zijun Wei$^1$} \\
  \textbf{Tianshu Yu$^1$ \quad
  Geoffrey Fox$^2$ \quad
  Peng Wu$^{1,\dagger}$ \quad
  Xiao Yu$^{1,\dagger}$} \\
  $^1$ByteDance \quad
  $^2$The University of Virginia \\
  $^*$Equal contribution \quad
  $^\dagger$Corresponding authors
}

\maketitle

\begin{abstract}
Modern LLM RL systems separate rollout generation from policy optimization. These two stages are expected to produce token probabilities that match exactly. However, implementation differences can make them assign different values to the same sequence under the same model weights, inducing Training–Inference Mismatch (TIM). TIM is difficult to inspect because it is entangled with off-policy drift and common stabilization mechanisms. In this work, we isolate TIM in a zero-mismatch diagnostic setting (\sysname), and show that small token-level numerical disagreements can independently cause training collapse. We further show that TIM changes the effective optimization problem, and identify a set of remedies that could mitigate TIM. Our results suggest that TIM is not benign numerical noise, but a systems-level perturbation that should be treated as a first-order factor in analyzing LLM RL stability.
\end{abstract}

\section{Introduction}

Large language model (LLM) reinforcement learning (RL) has become a central paradigm for post-training foundation models and a key driver of recent advances in complex reasoning capabilities~\citep{schulman2017ppo, ouyang2022training, ziegler2019fine, shao2024deepseekmath, stiennon2020learning, liu2025understanding}. However, RL training remains difficult to stabilize in practice: policies may rapidly degrade, causing reward signals to drop over short training windows. 

Understanding what causes these collapses is therefore essential for building reliable LLM RL training systems~\citep{verl,hu2024openrlhf,fu2025areal,miles2025,cao2025skyrl,sheng2025laminar, chen2025minimax}. However, diagnosing the root cause is difficult because many failure modes are deeply entangled. A collapse may be caused by many factors, such as poorly tuned hyperparameters, reward misspecification, and reward hacking~\citep{fu2025reward,pan2024feedback}.
Among these factors, Training-Inference Mismatch (TIM) is an infrastructure-level confounder: numerical differences between training and inference engines can cause divergent token probabilities, even for the same input and model weights.

In response to these stability challenges, the community has developed a range of training-level stabilization techniques, including importance sampling, rejection sampling, and other forms of conservative policy updates~\citep{schulman2017ppo, yao2025tis, li2026mis, lingteam2025icepop, zheng2025stabilizing, liu2025mis}. 
Although effective in some settings, their connection to specific failure mechanisms remains unclear: the same technique may correct PPO mini-step off-policy drift, suppress TIM-induced numerical outliers, or introduce additional optimization bias.
Without a TIM-free diagnostic baseline to isolate these effects, practitioners must tune interventions and filtering thresholds by trial and error rather than by causal diagnosis.

In this paper, we aim to systematically understand the impact of TIM on LLM RL stability. Specifically, we aim to answer two key questions: First, does TIM contribute to RL training instability, and if so, to what extent?
Second, how do common stabilization techniques interact with TIM, what aspects of the mismatch do they mitigate, and what optimization side effects do they introduce?

To answer these questions, we develop \sysname{}, a lightweight rollout engine that achieves zero-mismatch with FSDP~\citep{zhao2023fsdp, rajbhandari2020zeromemoryoptimizationstraining} engine on top of VeRL~\citep{verl}.
\sysname{} eliminates TIM by unifying kernel and model implementations with the FSDP training engine, and by employing batch-invariant kernels~\citep{he2025nondeterminism} (\S~\ref{subsec:sys}).
Using \sysname{}, we conduct fine-grained diagnostic studies of LLM RL stability. Concretely, our contributions are:

    \textbf{Isolating TIM impact for LLM RL:}
    Using our TIM-free baseline, we identify TIM alone as a significant factor in triggering RL training collapse (\S~\ref{subsec:moe_reinforce_diagnosis}).

    \textbf{Analyzing failure modes of TIM-induced RL training collapse.}
    We then conduct ablation studies on TIM's role in RL training collapse under general setups. 
    Specifically, we analyze why RL training collapse in both trainer-side log-probabilities recomputation and rollout-side log-probabilities bypass.
    We find that TIM fundamentally changes the optimization objective, thereby inducing distinct failure (\S~\ref{subsec:recomputation_bypass}).

    \textbf{Ablating effectiveness of algorithmic TIM compensation.}
    Furthermore, we evaluate whether common stabilization techniques can effectively mitigate TIM, including truncated importance sampling (TIS)~\citep{yao2025tis}, and rejection sampling (RS)~\citep{li2026mis}(\S~\ref{subsec:rollout_correction}).
    Based on our ablation study, we identify an effective combination of existing algorithmic TIM compensations that can closely track our TIM-free baseline.

\section{Training-Inference Mismatch in LLM RL}
\label{sec:tim_formulation}

Due to the implementation differences between the training engines (FSDP~\citep{zhao2023fsdp, rajbhandari2020zeromemoryoptimizationstraining}, Megatron~\citep{shoeybi2020megatronlmtrainingmultibillionparameter}, etc) and the inference engines (vLLM~\citep{vllm}, SGLang~\citep{zheng2024sglangefficientexecutionstructured}, etc), including divergent model/kernel implementations, the probability distribution on the vocabulary for the next token can be different even with the exact same model checkpoint and inputs. 
This introduces an unintended off-policy bias between the sampling and model update.
Different from the off-policy bias introduced by PPO mini-steps~\citep{schulman2017ppo}, TIM off-policy bias is an infrastructure-level noise, which cannot be addressed by naive PPO clipping methods (discussed in \S \ref{sec:analysis}).

We formulate this issue in RL objectives as follows: 
Given a context $x$ and a sampled response $y=(a_1,\ldots,a_T)$, let $s_t=(x,y_{<t})$. We distinguish three token-level distributions: $\pi_\theta(a_t|s_t)$ denotes the current policy being optimized; $\pi^{\mathrm{rollout}}_{old}(a_t|s_t)$ denotes the behavioral distribution realized by the rollout engine when the token is sampled; and $\pi^{\mathrm{train}}_{old}(a_t|s_t)$ denotes the trainer-side reference distribution used when an algorithm requires an old-policy probability.

In an exact on-policy implementation, the probability assigned to each sampled token should be consistent between rollout and training. TIM occurs when the rollout execution path and the trainer execution path assign different probabilities to the same token under the same model weights and sampled sequence. At the token level, this discrepancy can be written as
\begin{equation}
\delta_t
=
\log \pi^{\mathrm{train}}_{old}(a_t|s_t)
-
\log \pi^{\mathrm{rollout}}_{old}(a_t|s_t).
\end{equation}
This definition is objective-agnostic: the mismatch exists before choosing whether the update is implemented with REINFORCE~\citep{williams1992reinforce, hu2025reinforce_plus_plus}, PPO~\citep{schulman2017ppo}, or GRPO~\citep{shao2024deepseekmath, yu2025dapo}.

\begin{figure*}[t]
    \centering
    \begin{subfigure}[t]{0.35\textwidth}
        \centering
        \includegraphics[width=\linewidth]{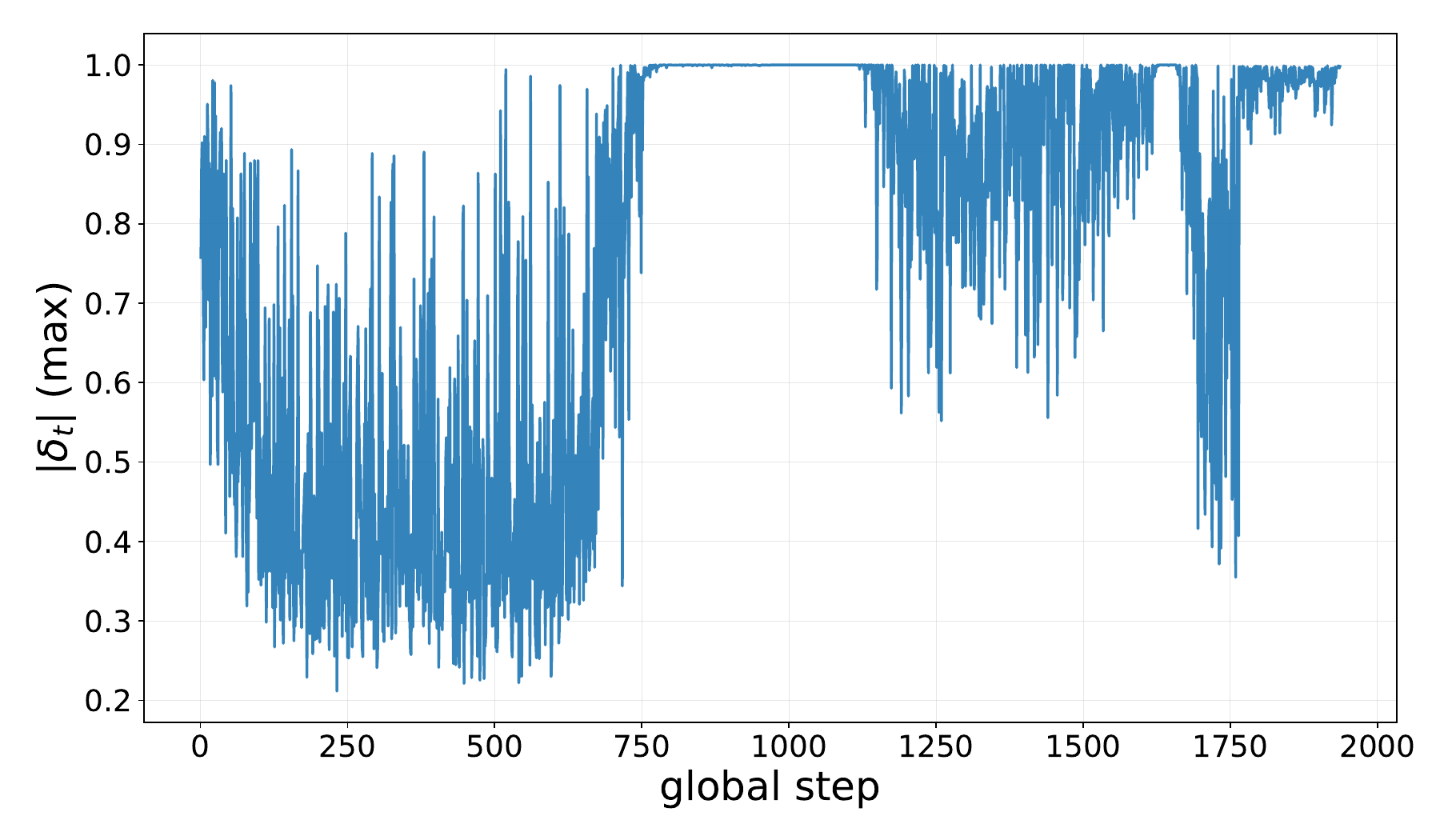}
        \caption{$|\delta_t|$ (max)}
    \end{subfigure}
    \begin{subfigure}[t]{0.35\textwidth}
        \centering
        \includegraphics[width=\linewidth]{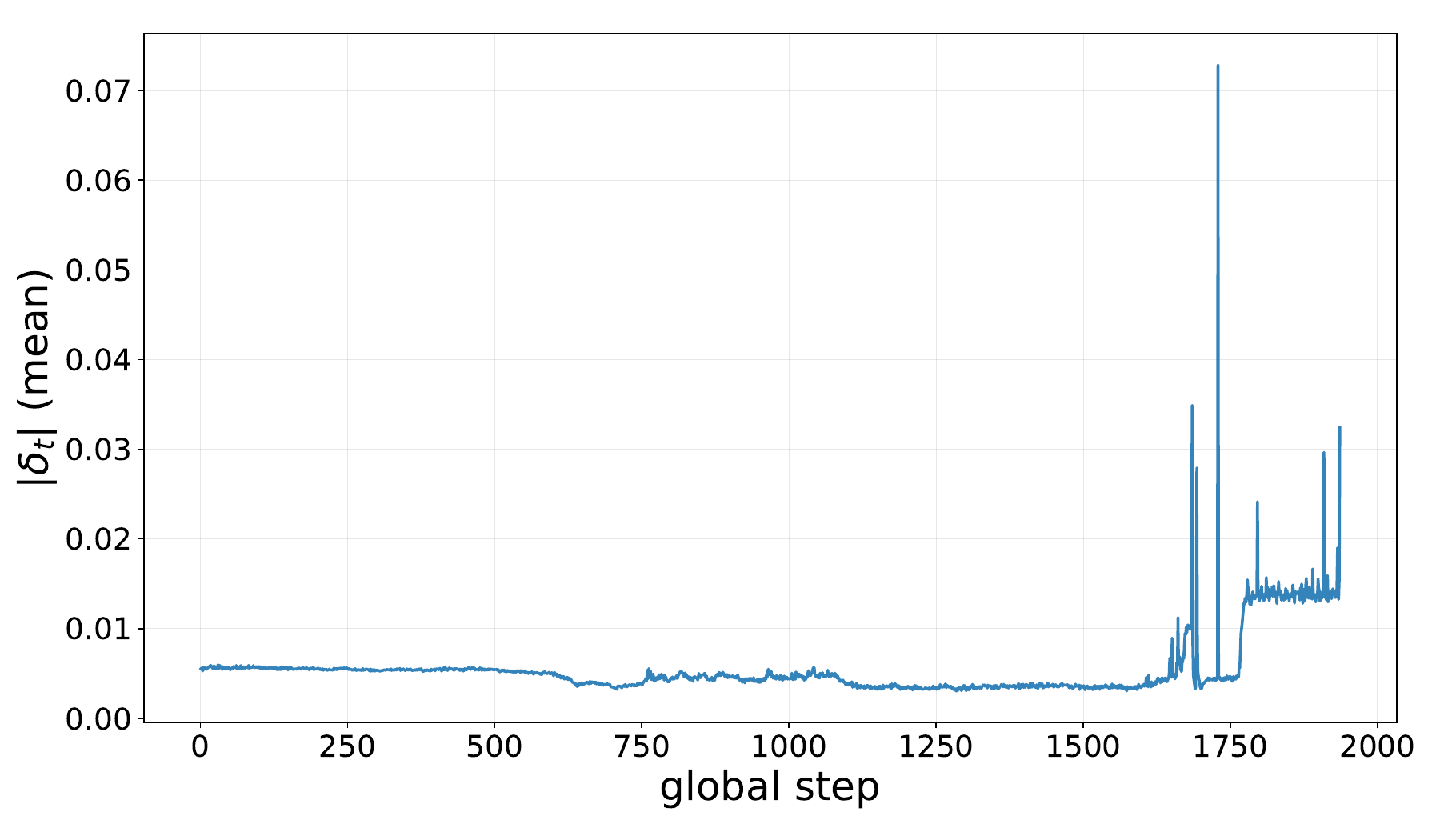}
        \caption{$|\delta_t|$ (mean)}
    \end{subfigure}

    \caption{Statistical $|{\delta_t}|$ max and mean for every training batch in the Qwen3-1.7B GRPO experiment (detailed configuration in Appendix~\ref{appendix:experimental_settings}). While the mean of $|\delta_t|$ is small, we can observe some extreme tokens with its $|{\delta_t}|$ near 1.0. 
    }
    \label{fig:delta_t_batch}
    
\end{figure*}

Table~\ref{tab:motivation-token-logprob} illustrates an example of several sampled tokens from the same context and model checkpoint. 
For each token, we compare the log-probability produced by the rollout with that re-evaluated by the training engine.
Ideally, these two values should be identical for an on-policy update. 
However, we can observe clear discrepancies in token log-probabilities between the two sides, including cases where the top-1 token choice at a given position is flipped.
As Figure~\ref{fig:delta_t_batch} shows, although the average difference in token-level probabilities is small per training batch, the maximum difference can even reach 1.0 for some extreme tokens, which is generally observable when TIM exists.

\definecolor{driftneg}{RGB}{215, 48, 39}   
\definecolor{driftpos}{RGB}{ 33,102,172}   
\begin{table}[t]
\centering
\caption{Token-level kernel-numerical drift between the rollout and training stacks on the same Qwen3-8B (bf16) weights, traced along one sentence from a greedy sampled response on an AIME-2024 problem.
$\delta_t = \log \pi_{\text{train}} - \log \pi_{\text{rollout}}$. Most positions are bit-close, but $^{\dagger}$ marks an argmax flip: at \texttt{that}, $\pi_{\text{train}}$'s top\,1 token is actually the punctuation\,+\,newline string \texttt{":\textbackslash n\textbackslash n"} (log\,prob $-0.577$), so the training side would have ended the clause as ``\textit{The problem states:}\textbackslash n\textbackslash n''.
}
\label{tab:motivation-token-logprob}

\scriptsize
\setlength{\tabcolsep}{3pt}
\begin{tabular}{l *{8}{r}}
\toprule
                            & \texttt{The} & \texttt{problem} & \texttt{states} & \texttt{that}$^{\dagger}$ & \texttt{there} & \texttt{exist} & \texttt{real} & \texttt{numbers} \\
\midrule
$\log \pi_{\text{rollout}}$ & $-0.279$ & $-0.063$ & $-0.314$ & $-0.694$ & $-0.000$ & $-0.030$ & $-0.000$ & $-0.000$ \\
$\log \pi_{\text{train}}$   & $-0.278$ & $-0.063$ & $-0.314$ & $-0.827$ & $-0.000$ & $-0.038$ & $-0.000$ & $-0.000$ \\
$\delta_t$             & \cellcolor{driftpos!2}  $\phantom{+}0.001$
                            & \cellcolor{white}       $\phantom{+}0.000$
                            & \cellcolor{white}       $\phantom{+}0.000$
                            & \cellcolor{driftneg!88} $-0.133$
                            & \cellcolor{white}       $\phantom{+}0.000$
                            & \cellcolor{driftneg!5}  $-0.008$
                            & \cellcolor{white}       $\phantom{+}0.000$
                            & \cellcolor{white}       $\phantom{+}0.000$ \\
\bottomrule
\end{tabular}

\end{table}

\section{Isolating Training-Inference Mismatch with \sysname{}}

\label{sec:tim_propagation}

In this section, we isolate the impact of TIM on RL training stability.
This requires two key ingredients: 
(1) First, we introduce a TIM-free rollout implementation, \sysname{}, as a diagnostic baseline that removes infrastructure-induced mismatch from the RL loop.
(2) Second, we evaluate this baseline under REINFORCE, which avoids PPO ratio clipping that may mask or distort TIM-induced changes in the loss and gradient signals.

\subsection{\sysname{}: A Zero-mismatch Rollout Engine}
\label{subsec:sys}

For the TIM-free baseline, we introduce \sysname{}, a lightweight rollout engine whose rollout token log-probabilities can achieve bit-wise alignment with the FSDP engine.

TIM comes from two sources:
(1) The model and kernel implementation differences between the inference and training engines.
Although semantically and mathematically the same, they often make different decisions regarding implementation details. 
For example, inference engines prefer inference-optimized kernel libraries like FlashInfer, which is not applicable in training engines.
(2) Variations in kernel reduction order and tiling.
Even when the same kernel implementation is used, performance-oriented optimizations such as atomic additions can introduce non-determinism, causing the kernel to produce different outputs for identical inputs. Moreover, even a \textit{deterministic} kernel may exhibit batch-dependent numerical behavior: changes in batch size can trigger different launch-grid configurations through auto-tuning, thereby altering GPU tiling strategies and reduction orders.
Since floating-point accumulation is non-associative under finite precision, these changes in execution order can ultimately lead to numerically different results.

Hence, \sysname addresses these two sources of mismatch by (1) using the same HuggingFace-based model implementation and register \sysname{} kernel implementation in the FSDP engine initialization and 
(2) employing deterministic and batch-invariant kernels, which fix the tiling and reduction order in the GPU kernel implementation.
Following the original batch invariant kernel implementation~\citep{he2025nondeterminism}, \sysname{} additionally implements RMSNorm~\citep{zhang2019rmsnorm}, batched matrix multiplication, and batch invariant Fused MoE kernels for efficient MoE training/inference.
For attention implementation~\citep{flashattention} specifically, we disable KV splitting~\citep{dao2023flashdecoding} to ensure determinism as well.

Meanwhile, since fixed tiling in batch-invariant kernels trades performance for numerical stability,
\sysname{} retains reasonable throughput for practical RL training by integrating chunked prefill~\citep{chunked_prefill}, CUDAGraph~\citep{nvidia2025cudagraphs}, pipeline parallelism~\citep{gpipe}, and optimistic KV allocation with preemption fallback. 
\sysname maintains hackable and very lightweight and its LOC is similar to nano-vLLM~\citep{nanovllm2025}
.

\begin{figure*}[t]
    \centering
    \begin{subfigure}[t]{0.24\textwidth}
        \centering
        \includegraphics[width=\linewidth]{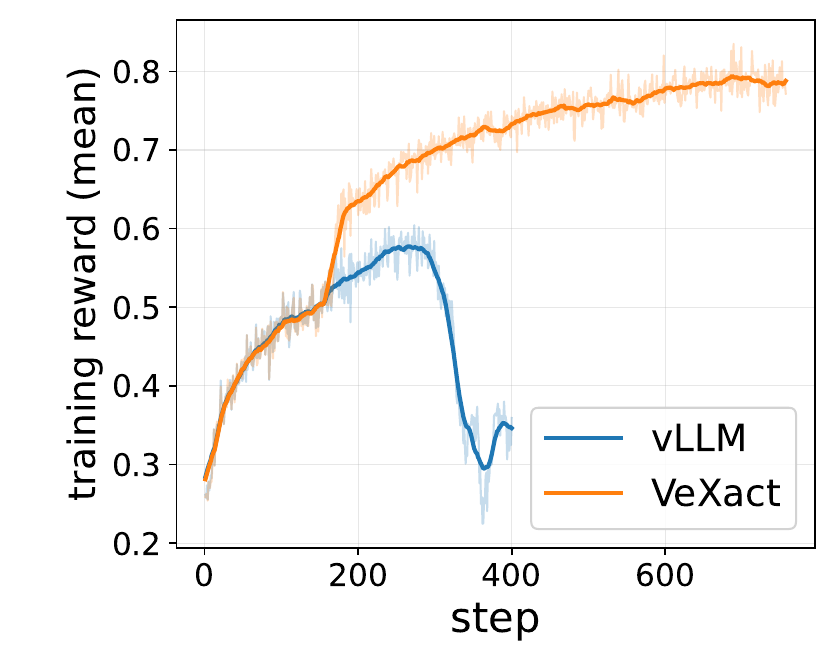}
        \caption{MoE training reward}
    \end{subfigure}
    \begin{subfigure}[t]{0.24\textwidth}
        \centering
        \includegraphics[width=\linewidth]{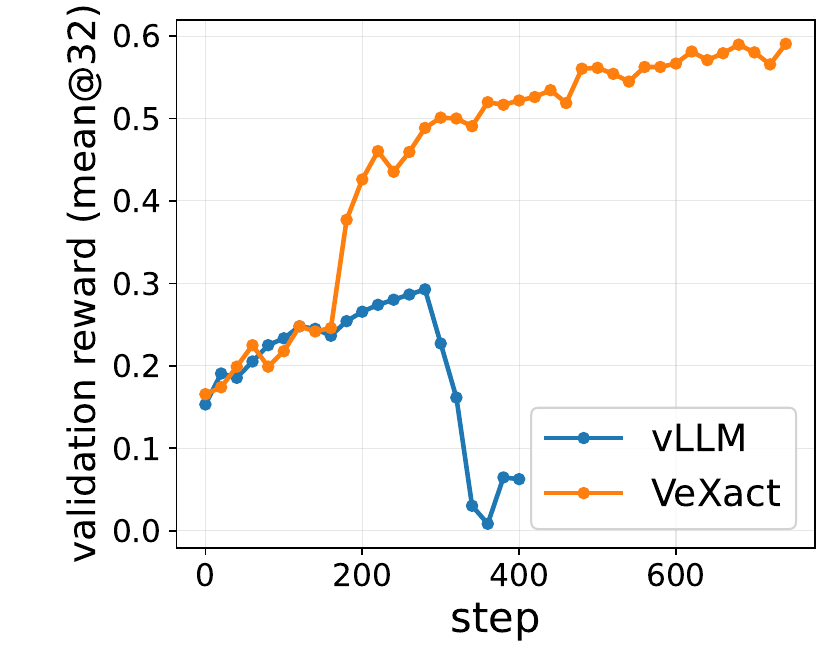}
        \caption{MoE val-reward}
    \end{subfigure}
    \begin{subfigure}[t]{0.24\textwidth}
        \centering
        \includegraphics[width=\linewidth]{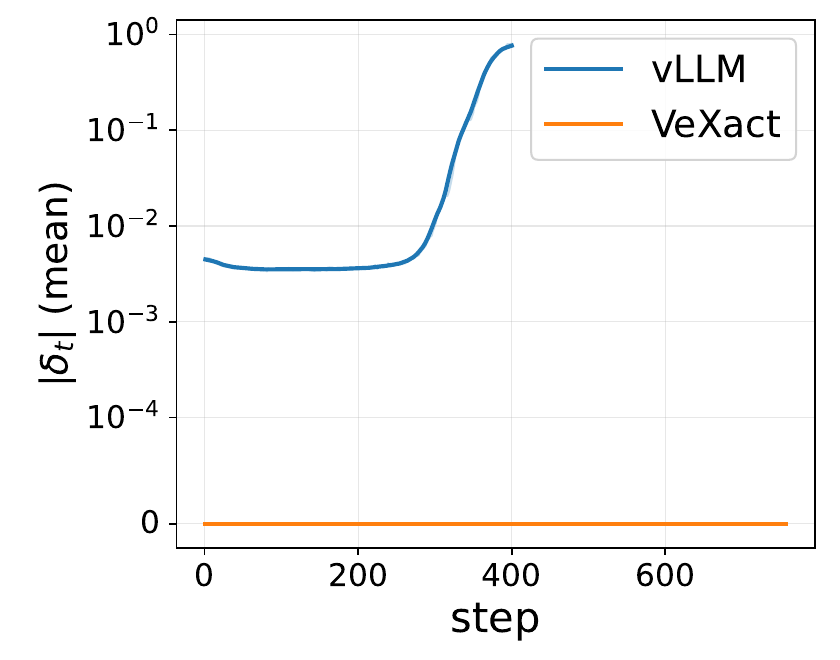}
        \caption{MoE $\delta_t$ (mean)}
    \end{subfigure}
    \begin{subfigure}[t]{0.24\textwidth}
        \centering
        \includegraphics[width=\linewidth]{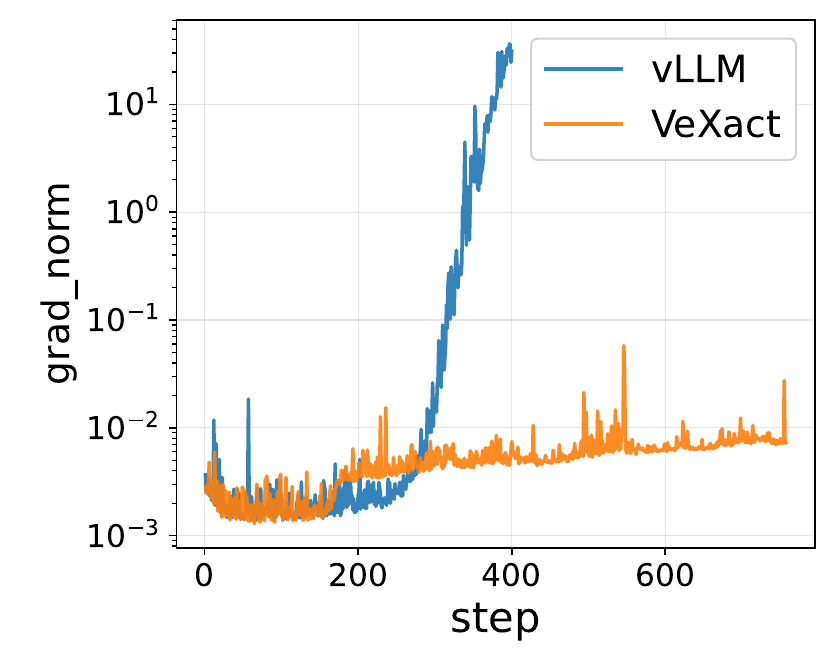}
        \caption{MoE gradient norm}
        \label{fig:reinforce-grad-moe}
    \end{subfigure}

    \centering
    \begin{subfigure}[t]{0.24\textwidth}
        \centering
        \includegraphics[width=\linewidth]{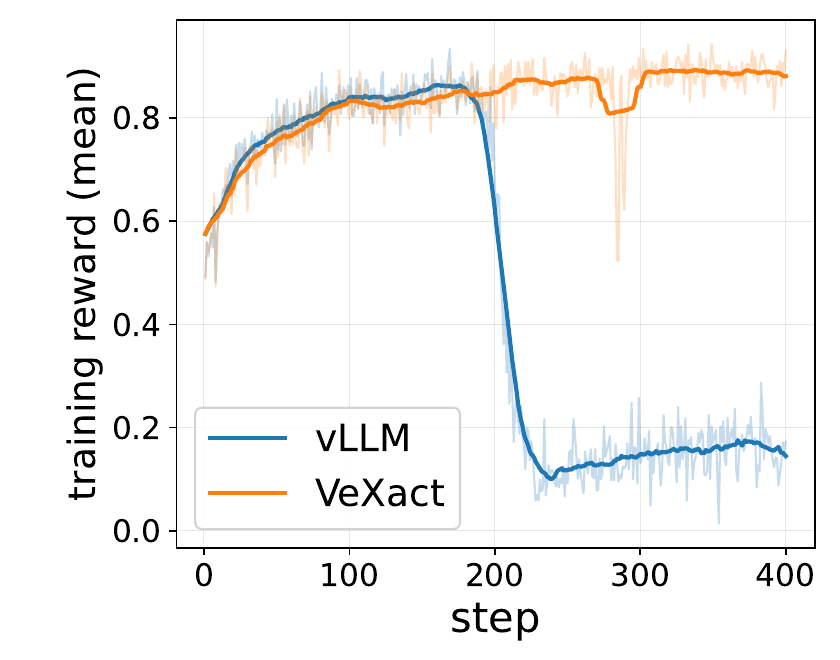}
        \caption{Dense training reward}
    \end{subfigure}
    \begin{subfigure}[t]{0.24\textwidth}
        \centering
        \includegraphics[width=\linewidth]{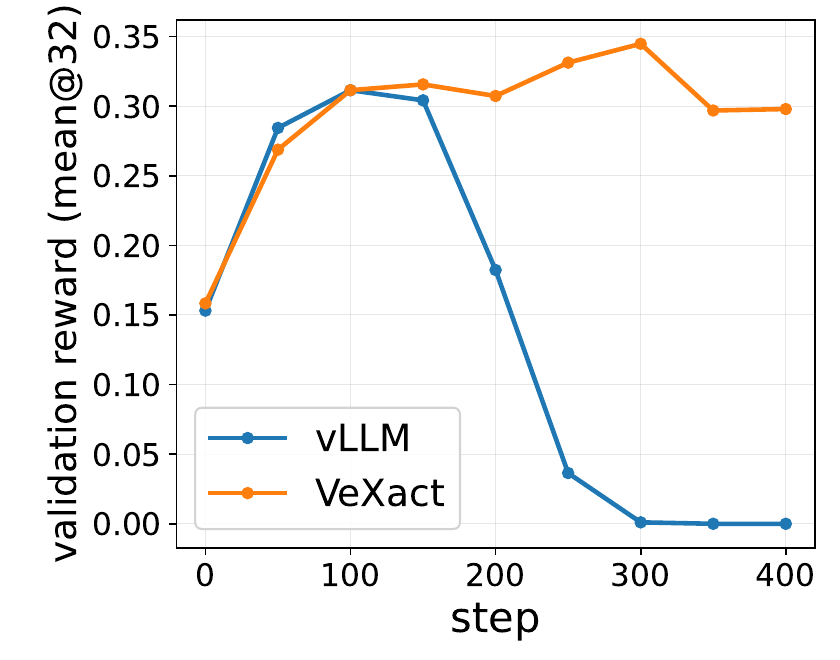}
        \caption{Dense val-reward}
    \end{subfigure}
    \begin{subfigure}[t]{0.24\textwidth}
        \centering
        \includegraphics[width=\linewidth]{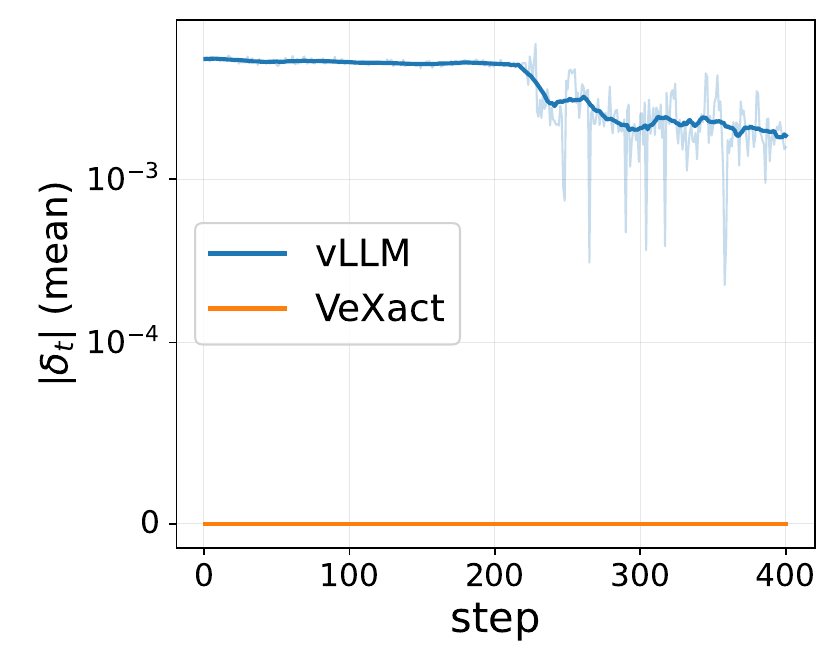}
        \caption{Dense $\delta_t$ (mean)}
    \end{subfigure}
    \begin{subfigure}[t]{0.24\textwidth}
        \centering
        \includegraphics[width=\linewidth]{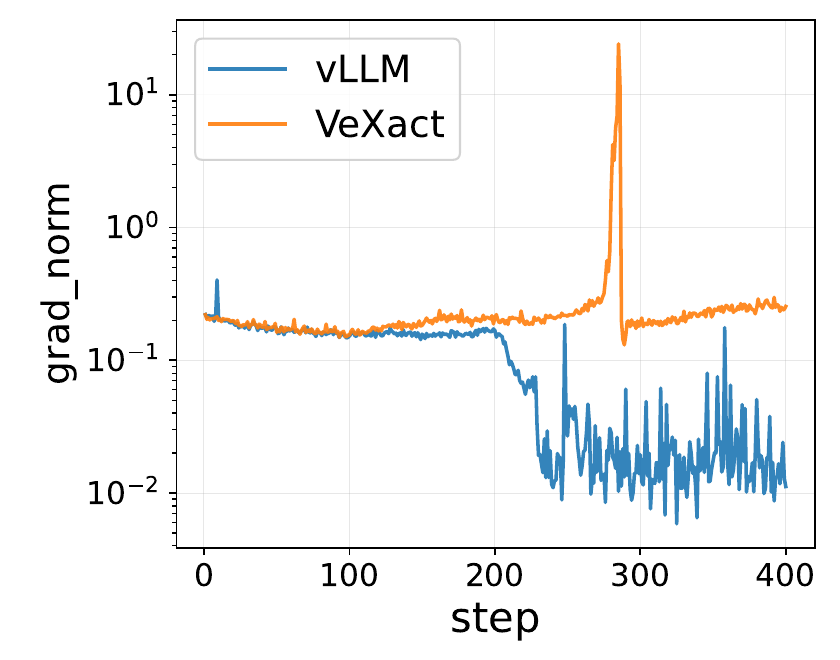}
        \caption{Dense gradient norm}
        \label{fig:reinforce-grad-dense}
    \end{subfigure}
    \caption{
    REINFORCE experiments comparing vLLM non-exact rollout with \sysname{}.
Top row: Qwen3-30B-A3B MoE. Bottom row: Qwen3-1.7B dense.
Each row reports training reward, AIME 2024 validation reward, $\delta_t$ (mean), and log-scale gradient norm. More results in Appendix~\ref{appendix:exp_reinforce}.
    }
    \label{fig:reinforce-reward-loss}
    
\end{figure*}

\subsection{Exposing the Impact of Training-Inference Mismatch}
\label{subsec:moe_reinforce_diagnosis}

To isolate how TIM impacts RL training stability from low-level numerical disagreement, we first study REINFORCE~\citep{williams1992reinforce} on-policy updates. Unlike PPO-style objectives, REINFORCE consumes each rollout batch in a single policy-gradient update. This makes it a cleaner diagnostic objective for attributing changes in loss, gradient, and reward to rollout-training probability disagreement.

We conduct REINFORCE experiments (with batch-whitened advantages) for both dense (Qwen3-1.7B) and MoE (Qwen3-30B-A3B) models~\citep{yang2025qwen3}. 
Both settings compare a standard non-exact rollout engine (vLLM) against \sysname{}.
The dense run is trained on Sanity-Test-R1D-1.5B~\citep{qi2025fp16} and evaluated on AIME 2024~\citep{aime2024} every 50 global steps. 
The MoE run is trained on DAPO~\citep{yu2025dapo} dataset and evaluated on AIME 2024 every 20 global steps. 
The detailed configuration is summarized in Appendix Table~\ref{tab:exp_setup}.

Figure~\ref{fig:reinforce-reward-loss} shows that the non-exact rollout run exhibits instability jointly in reward and gradient signals, while \sysname{} is significantly more stable.
For example, in the MoE setting of Figure~\ref{fig:reinforce-reward-loss}, the vLLM run initially improves but starts to degrade after step~$280$, with training and validation rewards decreasing from $0.574$ and $0.293$ to $0.255$ and $0.067$, respectively. By contrast, the \sysname{} reference remains stable and continues to improve, reaching $0.753$ training reward and $0.534$ validation reward. 
Since TIM is the only difference between \sysname{} and vLLM baseline,
\textit{these experiments confirm that TIM is itself a critical destabilizing factor, not merely a secondary artifact compounded with other training effects}.

\section{Ablating TIM Mitigations under the \sysname{}  Baseline}
\label{sec:analysis}

\S~\ref{sec:tim_propagation} shows that TIM alone can destabilize RL training with the REINFORCE setup.
However, algorithms like PPO and GRPO have more complicated setups with PPO mini-steps.
Different from fully on-policy algorithms like REINFORCE, algorithms like PPO and GRPO fix a snapshot of the policy ($\theta_{old}$) to collect a batch of samples and perform multiple gradient steps on this batch.
To optimize the current policy $\theta$ against the data sampled from a different distribution $\theta_{old}$, they apply \textit{importance sampling}, introducing the probability ratio  $\pi_{\theta}/\pi_{old}$ into the objective.

Under such setups, TIM is intertwined with stabilization techniques for them like PPO clipping. We now turn to the second research question: when practitioners apply common stabilization techniques, among which some are designed for general PPO training stability and some are TIM-aware, what do these mechanisms actually fix, and what optimization side effects do they introduce?

\noindent{\bf Experiment setup.} 
We study this question in a practical GRPO training setting. Unless otherwise stated, the experiments in this section use Qwen3-1.7B with FSDP on mathematical reasoning workloads. The model is trained on Sanity-Test-R1D-1.5B and evaluated on AIME 2024 every 50 global steps. We compare \sysname{} with vLLM non-exact rollout implementations using recomputation, bypass, rollout correction, and rejection sampling variants.

\begin{figure*}[t!]
    \centering
    \begin{subfigure}[t]{0.24\textwidth}
    \centering
    \includegraphics[width=0.99\linewidth]{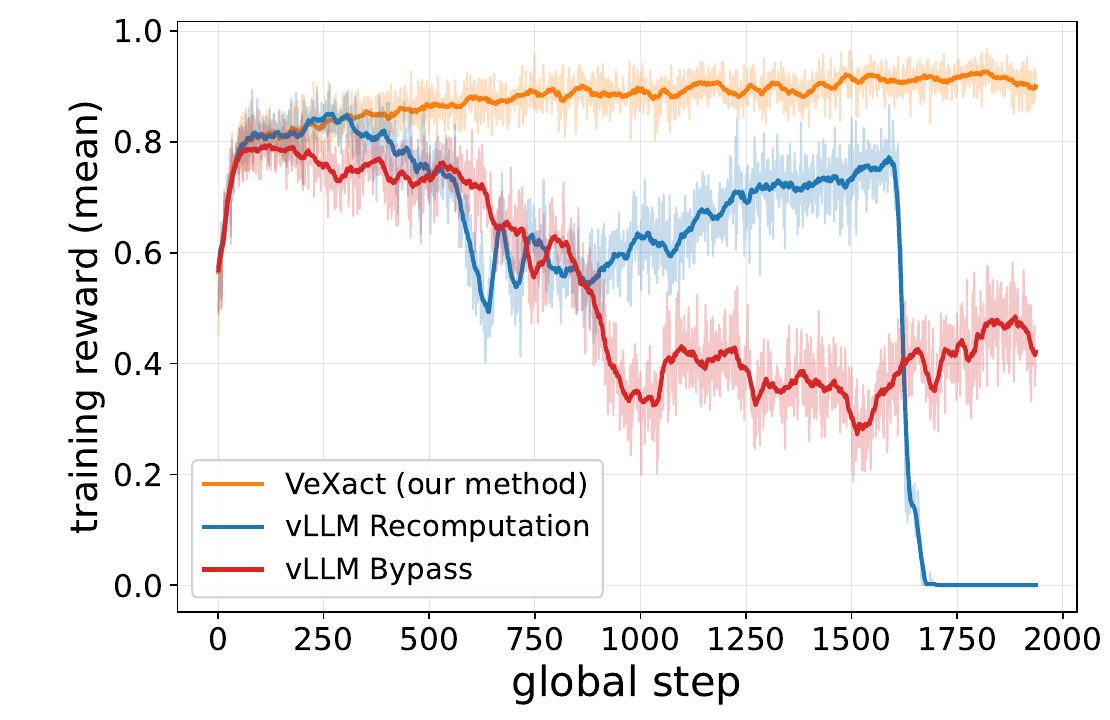}
    \caption{Training reward.}
    \end{subfigure}
    \begin{subfigure}[t]{0.24\textwidth}
    \centering
    \includegraphics[width=0.99\linewidth]{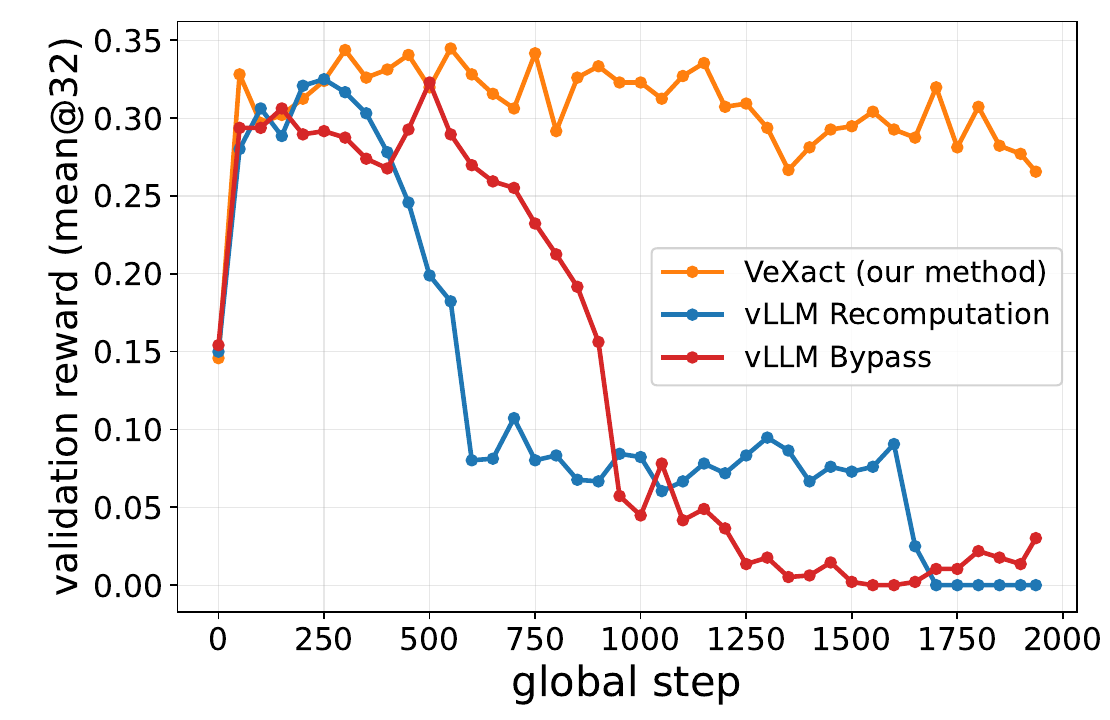}
    \caption{AIME24 val-reward.}
    \end{subfigure}
    \begin{subfigure}[t]{0.25\textwidth}
    \centering
    \includegraphics[width=0.99\linewidth]{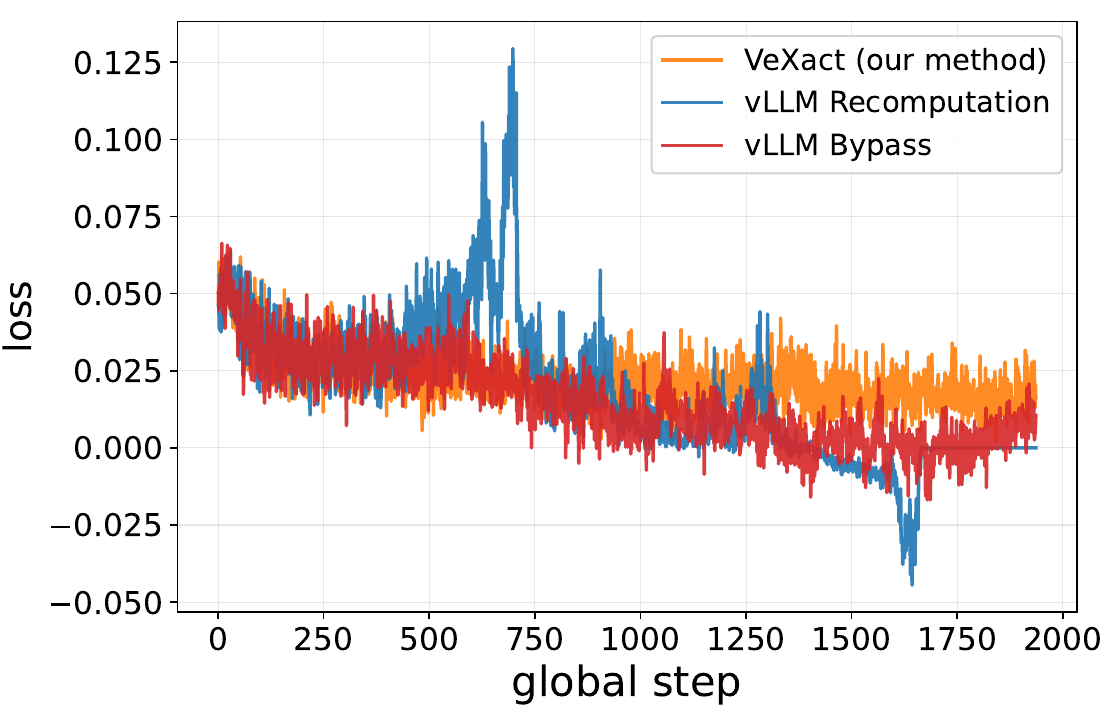}
    \caption{Loss.}
    \end{subfigure}
    \begin{subfigure}[t]{0.25\textwidth}
    \centering
    \includegraphics[width=0.99\linewidth]{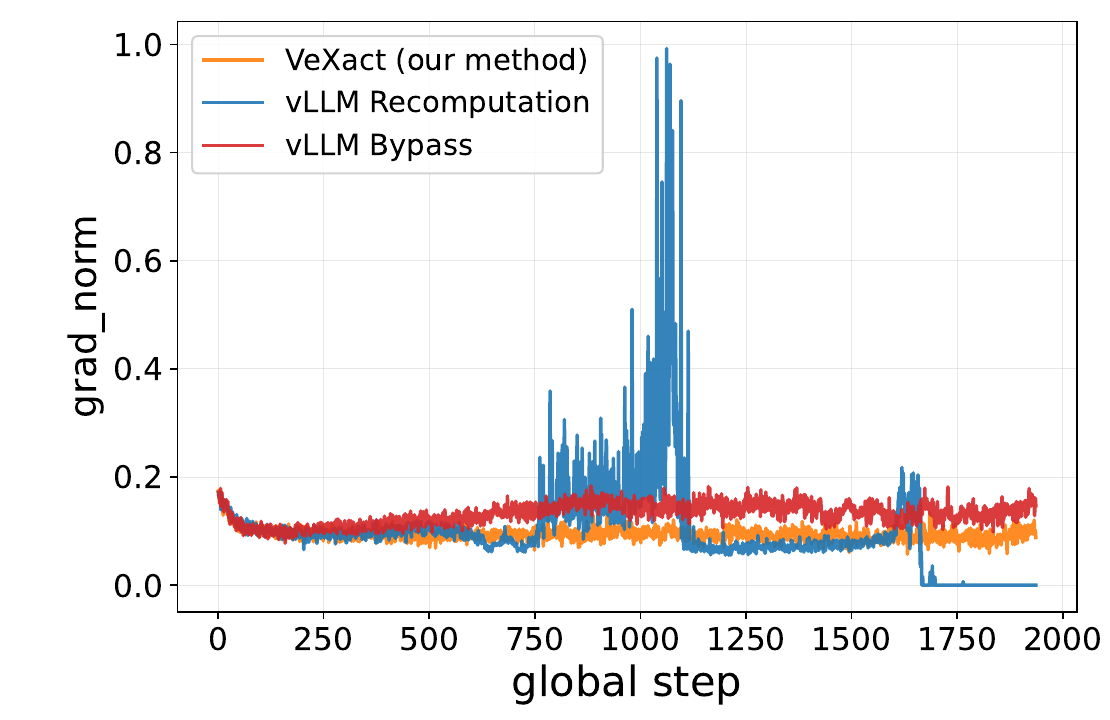}
    \caption{Gradient norm.}
    \end{subfigure}
    \caption{Qwen3-1.7B GRPO experiments with \sysname{} and vLLM recomputation and bypass, where only \sysname{} can maintain the training stability. Additional experimental results using the DAPO dataset, Llama models, and different random seeds are provided in Appendix~\ref{appendix:exp_dense_dapo}.}
    \label{fig:recompute_reward}
    
    \end{figure*}

\subsection{The Failure Modes of Recomputation and Bypass}
\label{subsec:recomputation_bypass}

There are two implementations for acquiring $\pi_{old}$: recomputation (where the trainer re-evaluates the sampled tokens) and bypass (where the rollout engine directly transmits its log-probabilities). 
Both strategies use the same clipped PPO/GRPO surrogate, but differ in how the denominator of the policy ratio is obtained.
For a sampled token $a_t$ under state $s_t$ with advantage $A_t$, the token-level clipped surrogate is
\begin{equation}
\mathcal{L}_{\mathrm{ppo}}(r_{ppo},A)
=
-\min\left(
r_{ppo}A,
\mathrm{clip}(r_{ppo},1-\epsilon,1+\epsilon)A
\right).
\label{eq:ppo_loss}
\end{equation}
Under recomputation and bypass, the PPO ratios are respectively defined as
\begin{equation}
r^{\mathrm{train}}_{ppo}
=
\frac{\pi_\theta(a_t\mid s_t)}
{\pi^{\mathrm{train}}_{old}(a_t\mid s_t)},
\qquad
r^{\mathrm{rollout}}_{ppo}
=
\frac{\pi_\theta(a_t\mid s_t)}
{\pi^{\mathrm{rollout}}_{old}(a_t\mid s_t)}.
\label{eq:ppo_ratio}
\end{equation}
Thus, recomputation and bypass instantiate the same clipped surrogate with different denominators:
\begin{equation}
\mathcal{L}_{\mathrm{recomp}}
=
\mathcal{L}_{\mathrm{ppo}}(r^{\mathrm{train}}_{ppo}, A),
\qquad
\mathcal{L}_{\mathrm{bypass}}
=
\mathcal{L}_{\mathrm{ppo}}(r^{\mathrm{rollout}}_{ppo}, A).
\end{equation}
The sequence-level training loss is obtained by summing these token-level terms over the response and averaging over the sampled batch.

Figure~\ref{fig:recompute_reward} shows the main phenomenon: while \sysname{} keeps the training reward around $0.93$, vLLM recomputation first degrades from about $0.87$ to roughly $0.40$ during the first $650$ steps, partially recovers afterward, and then drops rapidly again after about step $1610$ before collapsing to near-zero reward after about step $1665$.
vLLM bypass shows a single-stage degradation, with training reward dropping to roughly $0.4$ but not collapsing to zero, and this degradation is not accompanied by comparably large loss spikes. 
Unlike the REINFORCE setting in \S~\ref{subsec:moe_reinforce_diagnosis}, where reward collapse is accompanied by clear loss and gradient-norm anomalies, the GRPO runs in Figure~\ref{fig:recompute_reward} show less synchronized behavior across these signals: bypass exhibits reward degradation without a comparably synchronized loss anomaly, while recomputation enters an early reward-degradation phase before the later gradient-norm spike becomes visible. This motivates a finer-grained analysis.

\noindent{\bf KL estimators are not sufficient indicators.} Since TIM directly perturbs the effective distance between the updated policy $\pi_\theta$ and the old policy $\pi_{\mathrm{old}}$, we first inspect KL estimators computed on the PPO ratio~\citep{schulman2020kl}, such as $K_1(r_{\mathrm{ppo}}) = -\log r_{\mathrm{ppo}}$ and $K_3(r_{\mathrm{ppo}}) = (r_{\mathrm{ppo}} - 1) - \log r_{\mathrm{ppo}}$, where $r_{\mathrm{ppo}}$ can be instantiated as either $r^{\mathrm{train}}_{\mathrm{ppo}}$ or $r^{\mathrm{rollout}}_{\mathrm{ppo}}$ as defined in Eq.~\ref{eq:ppo_ratio}. In bypass mode, where the rollout and trainer probabilities become visibly inconsistent, these aggregate KL probes increase noticeably in both $K_1$ and $K_3$. However, under recomputation mode, even over the first 700 training steps in which the run already exhibits the onset of failure, the KL estimators remain close to the \sysname{} baseline and fail to expose the emerging instability.
This indicates that aggregate probability-space divergence is not sufficient to characterize the earliest stage of TIM-induced training failure.

\begin{figure*}[t!]
    \centering
    \begin{subfigure}[b]{0.4\textwidth}
        \centering
        \includegraphics[width=\linewidth]{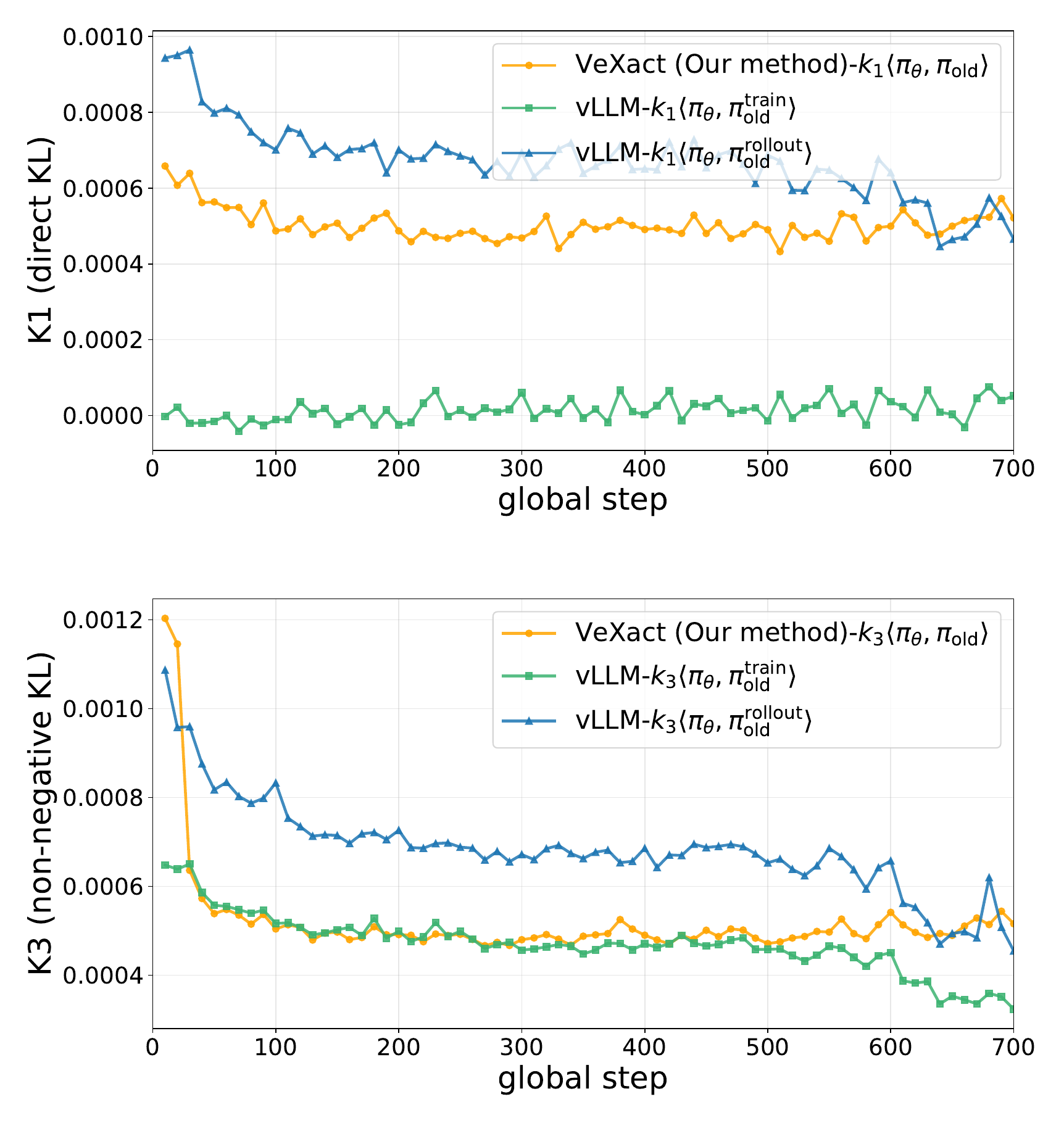}
        \caption{Recomputation: $K_1/K_3$ metrics}
        \label{fig:tim-stealth-k}
    \end{subfigure}
    \begin{subfigure}[b]{0.4\textwidth}
        \centering
        \includegraphics[width=\linewidth]{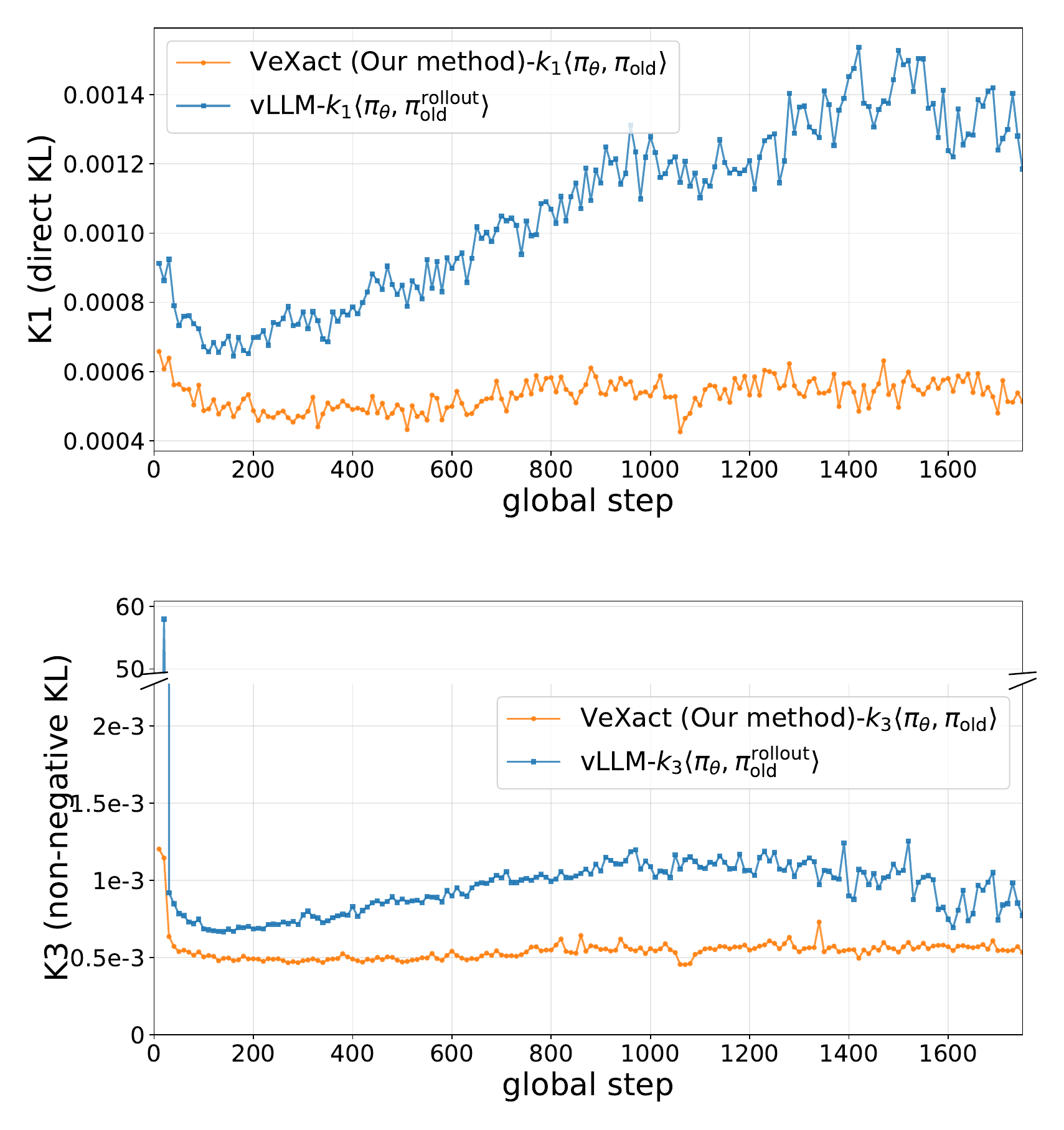}
        \caption{Bypass: $K_1/K_3$ metrics}
        \label{fig:tim-explosion}
    \end{subfigure}
    \caption{KL estimators under recomputation and bypass. In recomputation mode, both $K_1$ and $K_3$ remain nearly flat during the first 700 steps, even though the reward is already entering the degradation phase. In bypass mode, both $K_1$ and $K_3$ increase noticeably.  Under \sysname{}, $\pi_{\mathrm{old}}^{\mathrm{train}} = \pi_{\mathrm{old}}^{\mathrm{rollout}}$, so the two corresponding estimators coincide.
    With recomputation, $K_1\langle \pi_\theta, \pi_{\mathrm{old}}^{\mathrm{train}} \rangle$ is lower than \sysname{} because $K_1$ estimator is signed and each token's $K_1$ can cancel out when accumulating at the batch level.
    }
    \label{fig:recompute_kmetrics}
\end{figure*}

\noindent{\bf Zero-centered loss contribution.} 
The preceding analysis characterizes TIM in probability space. 
We further study how such probability space change impacts policy update.
First, considering that PPO loss applies different clipping boundaries to positive- and negative-advantage samples, we therefore analyze the two groups separately.
Second, we subtract $1$ from $r_{\mathrm{ppo}}$ to measure the loss change relative
to $r_{\mathrm{ppo}}=1$, removing the advantage-dependent offset $-A_t$.
For unclipped tokens, we define the zero-centered loss contribution:
\begin{equation}
C(r_{ppo})
=
-(r_{ppo}-1)A_t ,
\qquad
r_{ppo} \in \{r_{ppo}^{\mathrm{train}}, r_{ppo}^{\mathrm{rollout}}\},
\end{equation}

which has the same gradient as the standard $-r_{ppo}A_t$ objective but is zero when $r_{ppo}=1$. 
We evaluate this quantity using two denominators for the same sampled token. $r_{\mathrm{ppo}}^{\mathrm{train}}$ uses the trainer-recomputed distribution $\pi_{\mathrm{train}}$ and $r_{\mathrm{ppo}}^{\mathrm{rollout}}$ uses the rollout distribution $\pi_{\mathrm{rollout}}$. Comparing $C(r_{\mathrm{ppo}}^{\mathrm{train}})$ with $C(r_{\mathrm{ppo}}^{\mathrm{rollout}})$ therefore isolates how trainer-side recomputation alters the ratio-dependent surrogate contribution for the same sampled token. Under \sysname{}, the two distributions are numerically aligned, yielding $r_{\mathrm{ppo}}^{\mathrm{train}} = r_{\mathrm{ppo}}^{\mathrm{rollout}}$ and hence identical contributions.

\begin{figure*}[t!]
    \centering
    \begin{subfigure}[b]{0.32\textwidth}
        \centering
        \includegraphics[width=\linewidth]{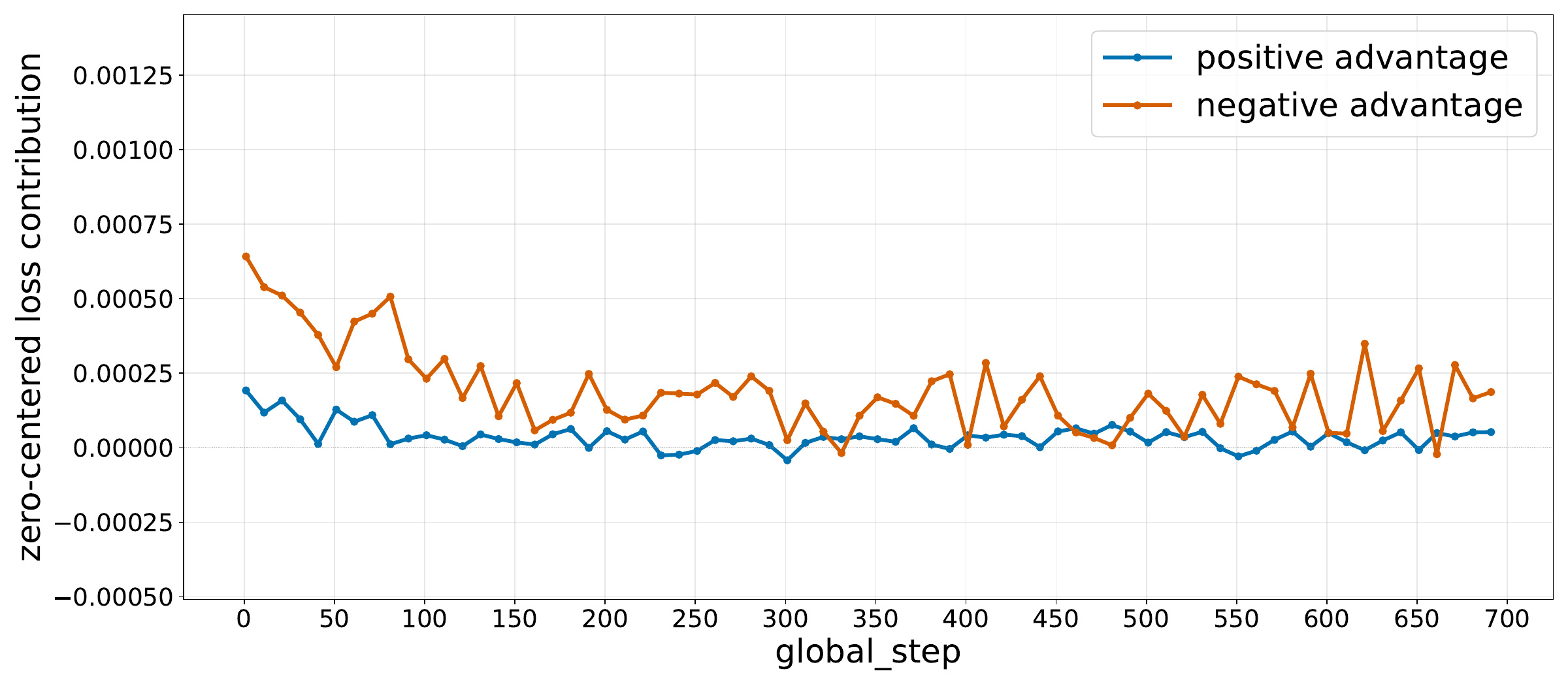}
        \caption{\sysname{} $C(r_{ppo}^{\mathrm{rollout}})$/$C(r_{ppo}^{\mathrm{train}})$}
    \end{subfigure}
    \begin{subfigure}[b]{0.32\textwidth}
        \centering
        \includegraphics[width=\linewidth]{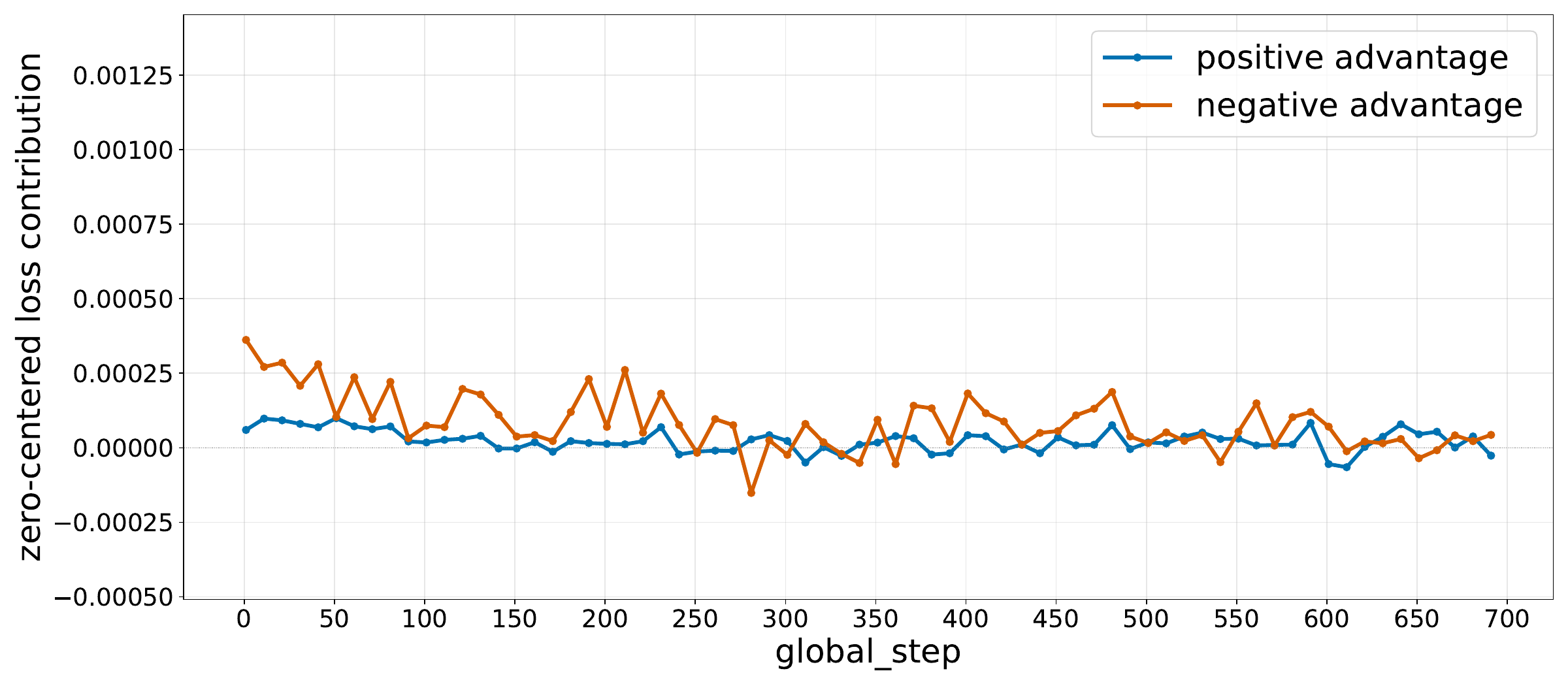}
        \caption{vLLM $C(r_{ppo}^{\mathrm{rollout}})$}
    \end{subfigure}
    \begin{subfigure}[b]{0.32\textwidth}
        \centering
        \includegraphics[width=\linewidth]{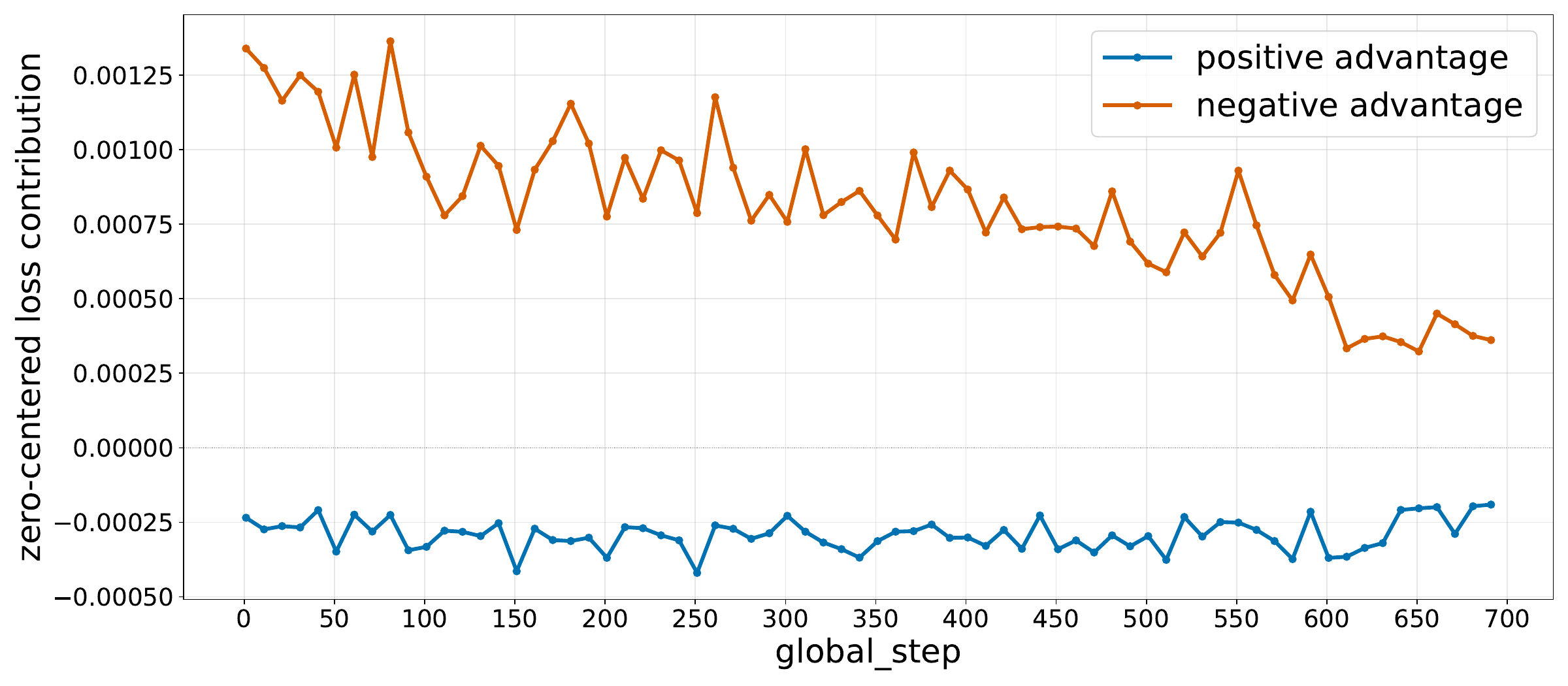}
        \caption{vLLM $C(r_{ppo}^{\mathrm{train}})$}
    \end{subfigure}
    \caption{
    Zero-centered loss contribution $C(r)=-(r-1)A$, which isolates the ratio-dependent component of the surrogate objective. 
    This shows that recomputation induces a sign-dependent skew in the advantage-weighted update signal, rather than merely adding uniform noise. 
}
    \label{fig:recompute_loss_dist}
    
\end{figure*}

Figure~\ref{fig:recompute_loss_dist} shows a clear sign-dependent distortion under TIM. With \sysname{}, rollout- and trainer-referenced contributions remain aligned. With vLLM, the rollout-referenced contribution stays near zero, whereas the trainer-referenced contribution separates by advantage sign, with a larger positive shift for negative-advantage tokens. Thus, even when aggregate KL estimators remain close to the \sysname{} baseline, recomputation can already introduce a systematic bias in the gradient-carrying update signal. Such biased updates can accumulate before the mismatch becomes visible as a clear KL spike.

\noindent{\bf Why bypass also fails.}
In bypass mode, the PPO ratio correctly uses the behavior distribution in the denominator (
$r^{\mathrm{rollout}}_{ppo}
=
{\pi^{train}_{\theta}}/
{\pi^{\mathrm{rollout}}_{old}}$
).
However, the numerator $\pi_\theta^{train}$ is evaluated using the trainer's numerical execution path.
During the backward pass, the optimizer computes the score function gradient based on $\nabla_\theta \log \pi_\theta^{train}$. Because TIM creates a misaligned probability landscape between the trainer and the rollout engine, the optimizer exploits numerical artifacts in the trainer's forward pass. These weight updates fail to translate into actual behavioral improvements when $\theta$ is deployed back to the rollout engine, leading to silent policy degradation. 
To validate this hypothesis, we conduct a minimal \textit{model update direction validation} experiment, which computes logprobs of past trajectories after the model update. 
For the tokens in the first 10 steps of training Qwen3-1.7B on the GSM-8K dataset, 17.3\% of token logprob change directions are misaligned with their advantage sign, while \sysname{} can ensure all the token logprob update directions are aligned. 

\paragraph{Takeaway.}
(1) In recomputation mode, the sampling log-probabilities used in loss computation are not from the actual samplers (rollout), resulting in a skew in the advantage-weighted loss contributions seen by the optimizer.
(2) Even with bypass mode, the actual optimization target still exists in a different sampling space from the rollout, making policy update ineffective.

\subsection{Ablating Algorithmic TIM Compensation}
\label{subsec:rollout_correction}

\S~\ref{subsec:recomputation_bypass} shows that recomputation and bypass instantiate different ways of accounting for the old-policy probability, but neither removes TIM at the source. This motivates a natural question: can post-hoc algorithmic compensation recover the behavior of zero-mismatch rollout, or does it only suppress some observable symptoms? We answer this question using \sysname{} as a TIM-free reference. Our goal is to diagnose what existing TIM-aware correction scaffolds can approximate, which design choices matter, and where their limits remain.

\paragraph{Existing algorithmic corrections.}
Existing rollout-correction methods commonly suppress unreliable samples through token-level truncation or sequence-level rejection. Let $r_{\mathrm{corr}} = \frac{\pi^{\mathrm{train}}_{\mathrm{old}}}{\pi^{\mathrm{rollout}}_{\mathrm{old}}}$ denote the correction ratio.
They typically act either at the token level truncated importance sampling,
\begin{equation}
\mathcal{L}_{\mathrm{TIS}}
=
\sum_{t=1}^{T}
\min(r_{\mathrm{corr},t}, \tau_{\mathrm{tok}})
\mathcal{L}_{\mathrm{PPO}}
\left(
r^{\mathrm{train}}_{\mathrm{ppo},t}, A_t
\right),
\end{equation}
or at the sequence level through rejection sampling,
\begin{equation}
\mathcal{L}_{\mathrm{RS}}
=
\mathbf{1}
\left[
S_{\mathrm{seq}}(q_{1:T}) \le \tau_{\mathrm{seq}}
\right]
\sum_{t=1}^{T}
\mathcal{L}_{\mathrm{PPO}}
\left(
r^{\mathrm{rollout}}_{\mathrm{ppo},t}, A_t
\right).
\end{equation}
Here, \(\tau_{\mathrm{tok}}\) is a token-level truncation threshold that limits large correction weights, and \(\tau_{\mathrm{seq}}\) is the sequence-level rejection threshold. For the specific threshold values, we follow prior work~\citep{qiu2026fp8rl,li2026mis}(more details in Appendix~\ref{app:patch_objectives}). \(q_{1:T}\) denotes the sequence of token-level diagnostic signals used for rejection, \(S_{\mathrm{seq}}(\cdot)\) maps these signals to a scalar trajectory-level trust-region score.

\paragraph{Our diagnostic instantiation.}
We therefore use these post-hoc corrections as diagnostic probes rather than as new algorithmic proposals.
\S~\ref{subsec:recomputation_bypass} shows that TIM can enter PPO/GRPO both through trainer--rollout old-policy displacement and through skewed token-level loss contributions.
These observations motivate a controlled ablation along two axes: the ratio used to drive sequence-level rejection and the effect of adding token-level truncation before sequence-level rejection.

For the masking signal, we compare
\(q_t \in \{ r_{\mathrm{corr},t}, r^{\mathrm{rollout}}_{\mathrm{ppo},t} \}\).
Here, \(r_{\mathrm{corr},t}\) measures system-induced mismatch between
\(\pi^{\mathrm{train}}_{\mathrm{old}}\) and \(\pi^{\mathrm{rollout}}_{\mathrm{old}}\),
whereas \(r^{\mathrm{rollout}}_{\mathrm{ppo},t}\) measures policy movement from the rollout behavior policy to the current policy.
For both choices, \(q_t\) is used only as the masking signal, rather than necessarily as the PPO ratio inside the surrogate objective.
At the granularity level, we apply sequence-level rejection to both signals by aggregating token-wise mismatch into
\(S_{\mathrm{seq}}(q_{1:T})=\sum_{t=1}^{T}K(q_t)\), with \(K(\cdot)\) instantiated as \(K_1\) or \(K_3\), and rejecting trajectories whose accumulated mismatch exceeds a threshold.
We evaluate a joint token- and sequence-level variant, which first filters localized token-level outliers and then applies the sequence-level rejection criterion.

Together, these choices yield three diagnostic configurations:
sequence-level rejection based on either \(r_{\mathrm{corr}}\) or \(r^{\mathrm{rollout}}_{\mathrm{ppo}}\), and joint token-level clipping plus sequence-level rejection based on \(r_{\mathrm{corr}}\), with the sequence-level score instantiated by either \(K1\) or \(K3\). 
 Objectives are detailed in Appendix~\ref{app:patch_objectives}.

\begin{figure*}[t]
    \centering
    \begin{subfigure}[t]{0.35\textwidth}
        \centering
        \includegraphics[width=\linewidth]{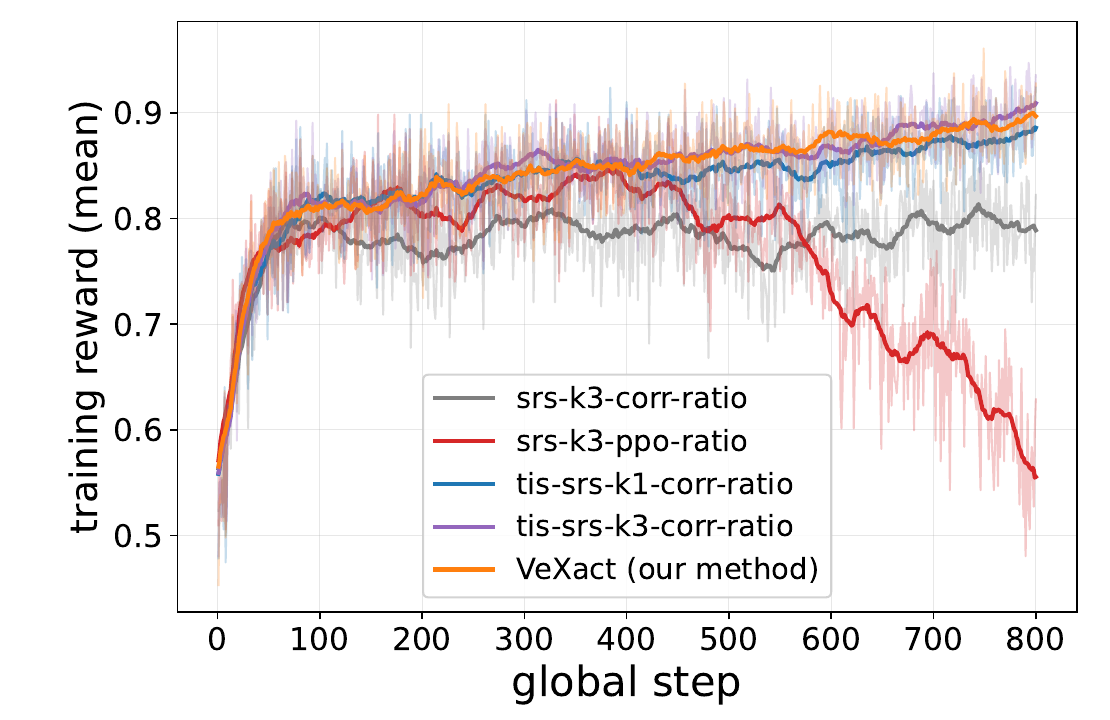}
        \caption{Training reward}
    \end{subfigure}
    \begin{subfigure}[t]{0.35\textwidth}
        \centering
        \includegraphics[width=\linewidth]{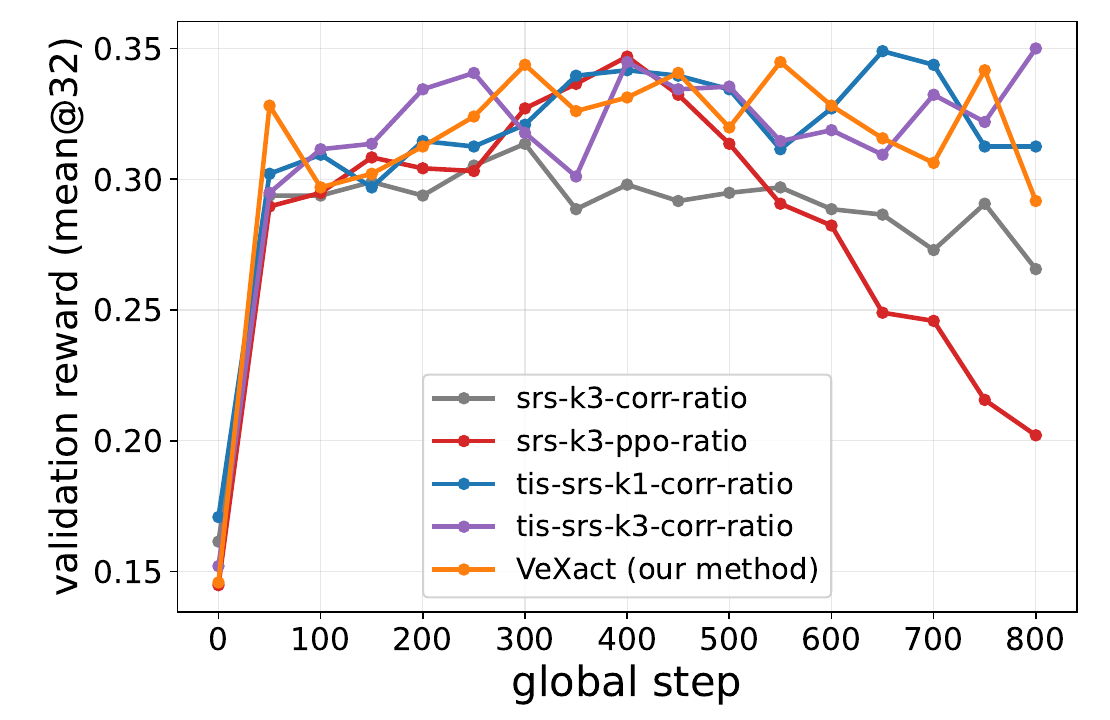}
        \caption{Validation reward}
    \end{subfigure}
    \begin{subfigure}[t]{0.35\textwidth}
        \centering
        \includegraphics[width=\linewidth]{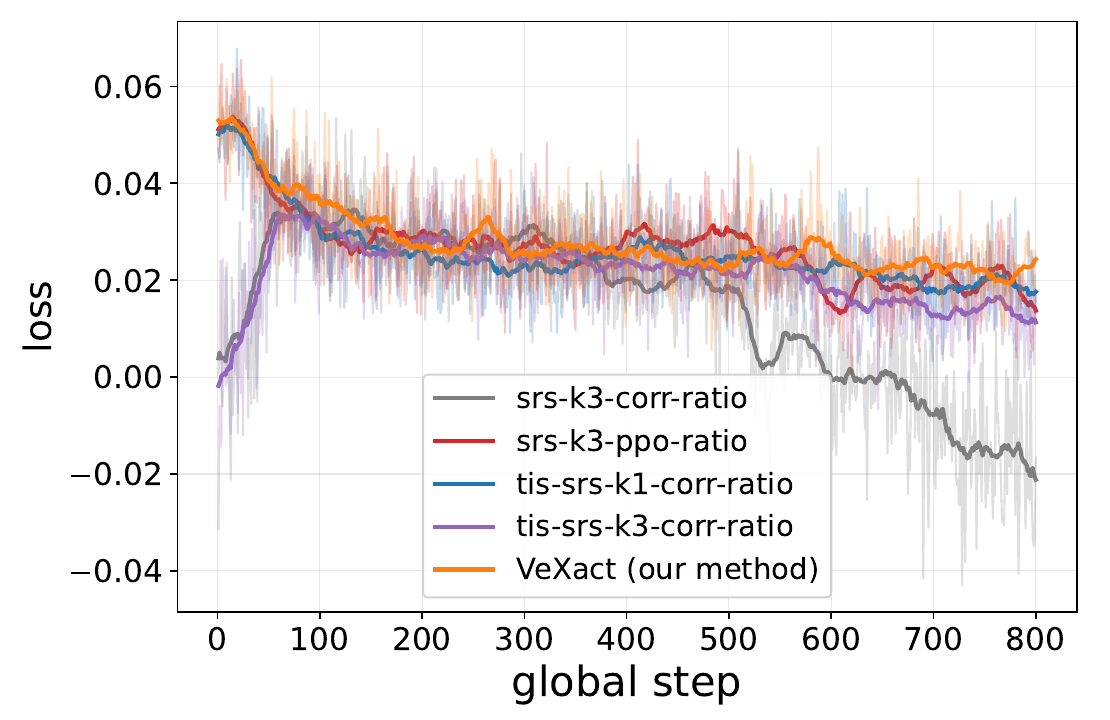}
        \caption{Loss}
    \end{subfigure}
    \begin{subfigure}[t]{0.35\textwidth}
        \centering
        \includegraphics[width=\linewidth]{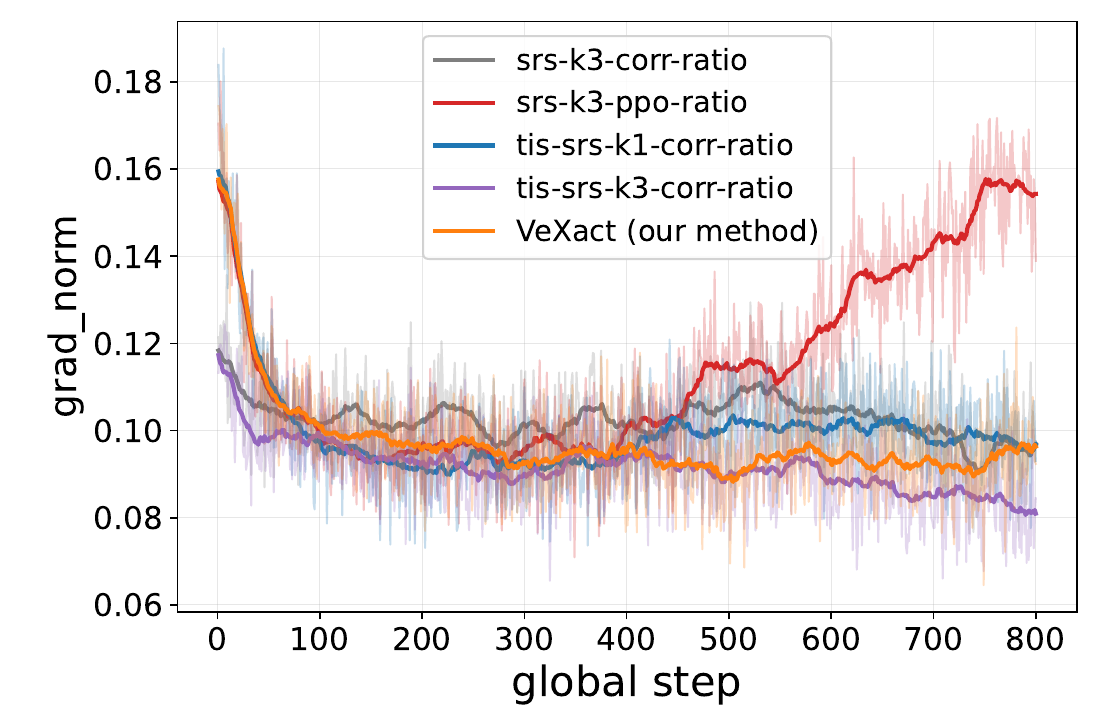}
        \caption{Gradient norm}
    \end{subfigure}
    \caption{Algorithmic-patch comparison. We evaluate four correction baselines: \texttt{srs-k3-corr-ratio} and \texttt{srs-k3-ppo-ratio} apply sequence-level rejection using $r_{\mathrm{corr}}$ and $r_{\mathrm{ppo}}$, respectively, while \texttt{tis-srs-k1-corr-ratio} and \texttt{tis-srs-k3-corr-ratio} combine truncated importance sampling with $r_{\mathrm{corr}}$-based sequence rejection instantiated with K1 and K3, respectively. We provide the expanded objectives in Appendix~\ref{app:patch_objectives}.
    For sequence rejection, $r_{corr}$ is more effective than $r_{ppo}$.
    Additionally with TIS, $r_{corr}$-based sequence rejection can track \sysname{} closely.
    For better clarity, curves are smoothed by center mean with window size of 25.
    }
    \label{fig:patch_ablation}
    
\end{figure*}

\noindent{\bf Experiment results.}
As Figure~\ref{fig:patch_ablation} shows,
comparing the sequence-level variants, using $r_t^{\mathrm{corr}}$ as the filtering signal yields higher reward than using $r^{\mathrm{rollout}}_{\mathrm{ppo},t}$. 
We attribute this difference to the distinct semantics of the two ratios:
The correction ratio clipping guarantees that the \textit{base distribution ($\pi_{old}^{train}$) to update is not too far from the sampling space.}
In contrast, the rollout ratio overlaps with PPO's policy-ratio mechanism, whose purpose is to control the update magnitude of how far the current policy moves away from the behavior policy, but not the starting distribution location.

The same ablation also shows that the configuration \textit{combining token-level filtering with sequence-level rejection} most closely tracks the \sysname{} reference, indicating that TIM manifests at multiple granularities. Localized token-level mismatch outliers can distort individual PPO contributions even when the aggregate sequence-level score remains moderate, while some trajectories exhibit large accumulated mismatch and should be rejected at the sequence level. 
The choice between \(K1\) and \(K3\) for the sequence rejection has a comparatively minor effect in our experiments. 

These results suggest that algorithmic correction can closely approach the zero-mismatch reference in the evaluated setting. However, these forms of algorithmic TIM compensation remains post-hoc: unlike \sysname{}, it can only suppress already-generated samples and may also discard useful learning signals.
More importantly, without \sysname{} as a ground-truth reference, patch configuration (e.g., $\tau_{\mathrm{seq}}$, $\tau_{\mathrm{tok}}$) remains largely ad hoc. \sysname{} therefore complements algorithmic patches by enabling their principled calibration.

\textbf{Takeaway.}
(1) $r_{corr}$-based sequence-level rejection sampling is more effective than $r_{ppo}$-based. The threshold metric choice of $K_1$ or $K_3$ does not affect much. 
(2) TIS is effective as it fixes the PPO ratio used in loss function from $r_{ppo}^{train}$ to $r_{ppo}^{rollout}$ by multiplying $r_{ppo}^{train}$ with $r_{corr}$, consistent with our conclusion from \S~\ref{subsec:recomputation_bypass}.
(3) Combining (1) and (2), we find that carefully designed algorithmic configurations can closely track our zero-mismatch reference \sysname.

\section{Related Works}

\noindent{\bf RL stabilization techniques.}
Existing stabilization techniques for RL training fall into
three categories.
\textit{Fine-grained clipping regions} such as DAPO~\citep{yu2025dapo} apply a non-symmetric clipping region for tokens with positive and negative advantages.
GSPO~\citep{zheng2025gspo} moves to sequence-level importance ratios to reduce variance accumulation over long responses;
\textit{TIM-aware corrections} explicitly target the training–inference numerical mismatch: 
Truncated IS (TIS)~\citep{yao2025tis} truncates extreme token-level IS weights, and Masked IS~\citep{zheng2025stabilizing, li2026mis} suppresses highly divergent samples by zeroing gradient contributions at either the token or sequence level. Score-centering~\citep{marek2026scorecenteringstabilizesoffpolicy} decomposes policy update gradients into a drift term caused by training-inference mismatch and a reward-covariance signal term, and removes the drift by subtracting the sampler's expected score, stabilizing off-policy RL without importance ratios.
Additionally, \citep{qi2025fp16} and \citep{dirhoussi2026_defeating_the_trainer_generator_precision_mismatch_in_trl} find that using FP16 instead of BF16 is helpful for RL training stability.
\textit{MoE-specific corrections} address the additional instability from dynamic expert routing.
For example, R3~\citep{ma2025stabilizing} directly replays inference-time routing decisions during training mini-steps to eliminate divergent expert selections.
Unlike these TIM compensation mechanisms, \sysname{} eliminates the root cause of the numerical mismatch.

\noindent{\bf Efficient batch-invariant kernels.}
To address the performance penalty of fixing tiling and reduction order in batch invariant kernels, DeepSeek-V4~\citep{deepseekai2026deepseekv4} introduces dual-kernel strategies for the attention kernel and a set of optimizations for GEMM kernels.
Tree-based invariant kernels (TBIK)~\citep{tbik} resolve the accumulation order mismatch across tensor-parallel orders by using a tree-based reduction order, enabling zero mismatch rollout under TP inference.
They prove that batch-invariant kernels can be highly performant and scalable for large-scale RL.
Our work focuses more on the analysis of TIM's role in RL training collapse with \sysname{} baseline.

\section{Discussion and Limitations}
\label{sec:discussion}

\noindent{\bf Relationship with asynchronous LLM RL.}
Large-scale agentic RL setups are typically asynchronous to speed up training, which introduces policy staleness as another off-policy source. 
Concurrent work~\citep{marek2026scorecenteringstabilizesoffpolicy} finds that corrections effective against policy staleness (e.g., PPO clipping) fail under numerical mismatch, indicating that the two are distinct sources of off-policy behavior that co-occur in asynchronous RL pipelines.
\sysname{} removes the numerical mismatch, allowing TIM to be isolated from staleness. 
Moreover, \sysname{} allows practitioners to systematically tune algorithmic setups and sensitive filtering thresholds to address the off-policy issues introduced by staleness in a noise-free environment.

\noindent{\bf Limitations.}
For the convenience of comparison with algorithmic compensation methods, our experiments primarily focus on the math datasets, as these works also mainly report results on math datasets.
We will evaluate broader tasks like coding in the future work.

\section{Conclusion}

We investigate Training-Inference Mismatch (TIM) as a systems-level confounder in LLM RL stability. With our zero-mismatch baseline \sysname{}, we demonstrate that TIM alone can destabilize RL training under a wide range of setups.
We further analyze how TIM alters the effective optimization problem, explaining why common implementation choices such as trainer-side recomputation and rollout-side bypass fail to fully eliminate its impact. 
Using \sysname{} as a reference, we show that carefully configured combinations of existing algorithmic corrections can closely match its performance.
Overall, our findings call for a joint system-algorithm perspective on RL stability and highlight the need for zero-mismatch RL execution.


\renewcommand{\acksection}{\section*{Acknowledgments}}
\begin{ack}
We are grateful to Haolin Liu (University of Virginia), Yuxuan Tong, Xibin Wu and Xia Xiao for their valuable suggestions and insightful discussions.
\end{ack}

\bibliographystyle{plainnat}
\bibliography{references}

\appendix
\newpage
\section{Additional Details}

\subsection{Experimental Settings}
\label{appendix:experimental_settings}

\begin{table}[htb]
\centering
\caption{Experimental settings corresponding to the dense GRPO, dense REINFORCE, and MoE REINFORCE recipes.}
\label{tab:exp_setup}

\small
\begin{tabular}{@{}  >{\raggedright\arraybackslash}p{0.16\linewidth}  >{\raggedright\arraybackslash}p{0.24\linewidth}  >{\raggedright\arraybackslash}p{0.24\linewidth}  >{\raggedright\arraybackslash}p{0.24\linewidth}@{}}
\toprule
\textbf{Setting} & \textbf{Dense GRPO} & \textbf{Dense REINFORCE} & \textbf{MoE REINFORCE} \\
\midrule
Model & Qwen3-1.7B, Llama-3.2-1B & Qwen3-1.7B & Qwen3-30B-A3B \\
RL algorithm & GRPO & REINFORCE & REINFORCE \\
Training data & Sanity-Test-R1D-1.5B & Sanity-Test-R1D-1.5B & DAPO-Math-17k \\
Evaluation & AIME 2024 every 50 steps & AIME 2024 every 50 steps & AIME 2024 every 20 steps \\
Batching & Global batch 64, mini-batch 16, rollout group 8 & Global batch 64 & Global batch 512 \\
Sequence length & Prompt 1024, response 8192 & Prompt 1024, response 8192 & Prompt 2048, response 20480 \\
Hardware & 1 node, 8 GPUs (H100) & 2 nodes, 16 GPUs (H100) & 8 nodes, 64 GPUs (H100) \\
Engine & FSDP2 + vLLM/\sysname{} & FSDP2 + vLLM/\sysname{} & FSDP2 + vLLM/\sysname{} \\
\bottomrule
\end{tabular}

\end{table}

\subsection{Additional REINFORCE Results}
\label{appendix:exp_reinforce}
\begin{figure}[htb]
    \centering
    \begin{subfigure}[t]{0.34\textwidth}
        \centering
        \includegraphics[width=\linewidth]{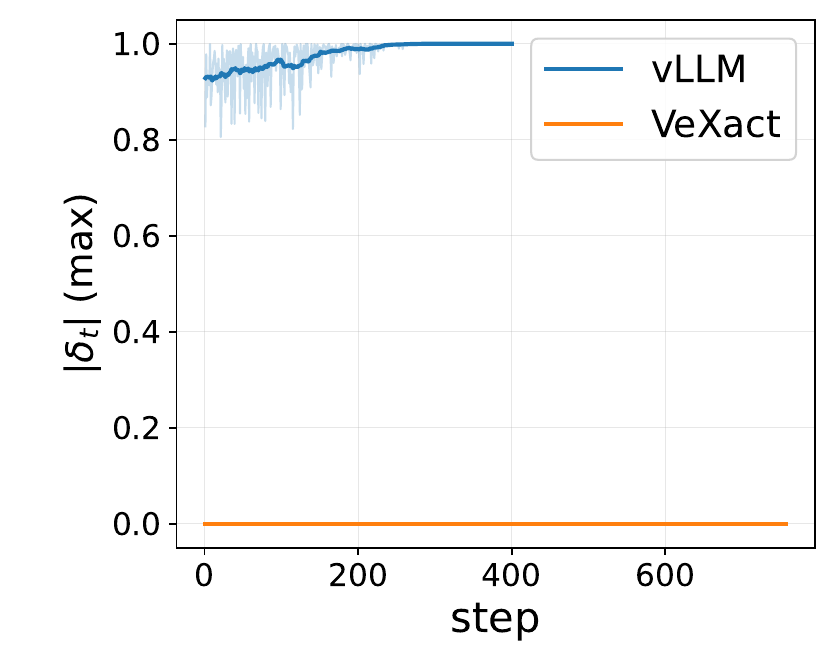}
        \caption{MoE $\delta_t$ (max)}
    \end{subfigure}
    \begin{subfigure}[t]{0.34\textwidth}
        \centering
        \includegraphics[width=\linewidth]{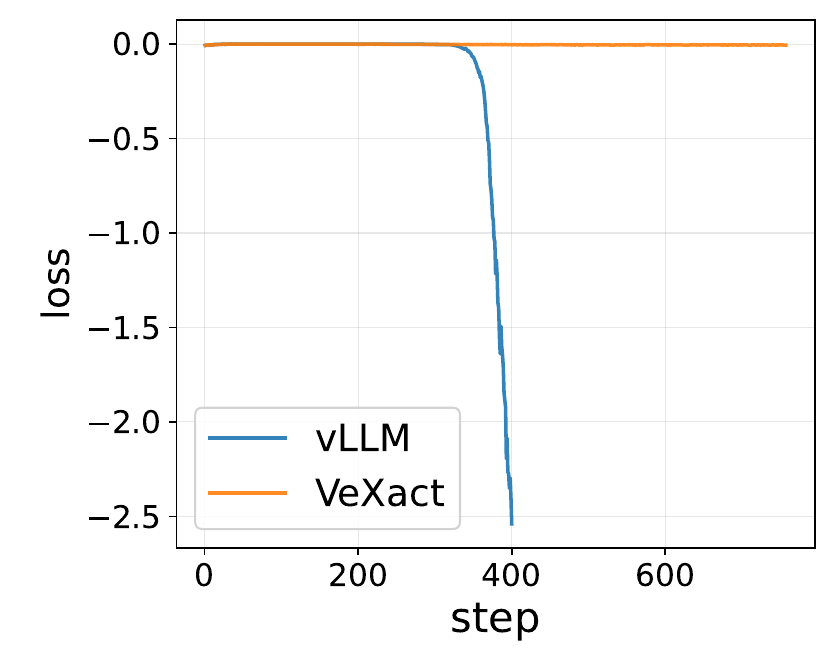}
        \caption{MoE sequence loss}
    \end{subfigure}

    \centering
    \begin{subfigure}[t]{0.34\textwidth}
        \centering
        \includegraphics[width=\linewidth]{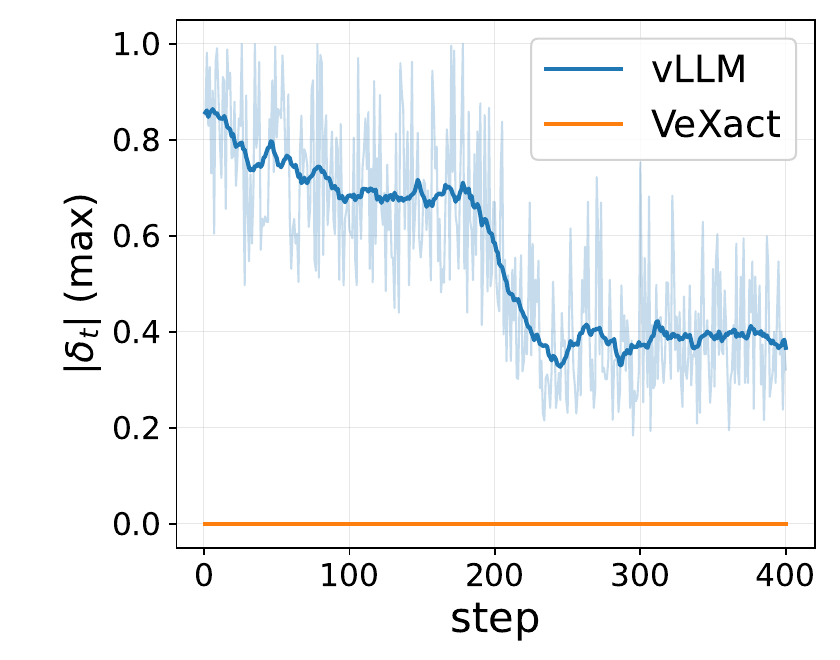}
        \caption{Dense $\delta_t$ (max)}
    \end{subfigure}
    \begin{subfigure}[t]{0.34\textwidth}
        \centering
        \includegraphics[width=\linewidth]{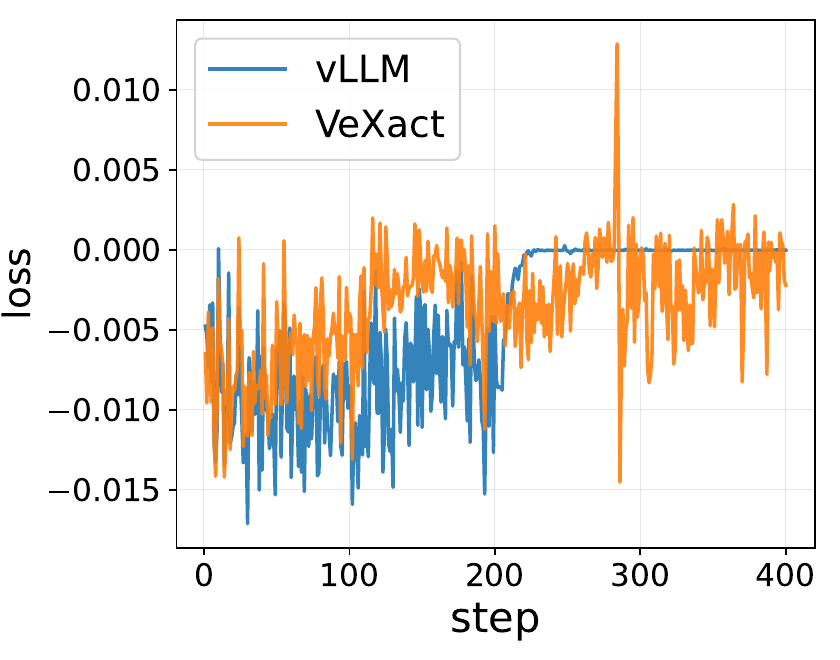}
        \caption{Dense sequence loss}
    \end{subfigure}
    \caption{
    Additional metrics of REINFORCE experiments comparing vLLM non-exact rollout with \sysname{}.
Each row reports $\delta_t$ (max), and sequence loss.
Top row: Qwen3-30B-A3B MoE. Bottom row: Qwen3-1.7B dense.
    }
\end{figure}

\newpage
\subsection{Additional Experiment Results}
\label{appendix:exp_dense_dapo}
\begin{figure}[htb]
    \centering
    \begin{subfigure}[t]{0.34\textwidth}
    \centering
    \includegraphics[width=0.99\linewidth]{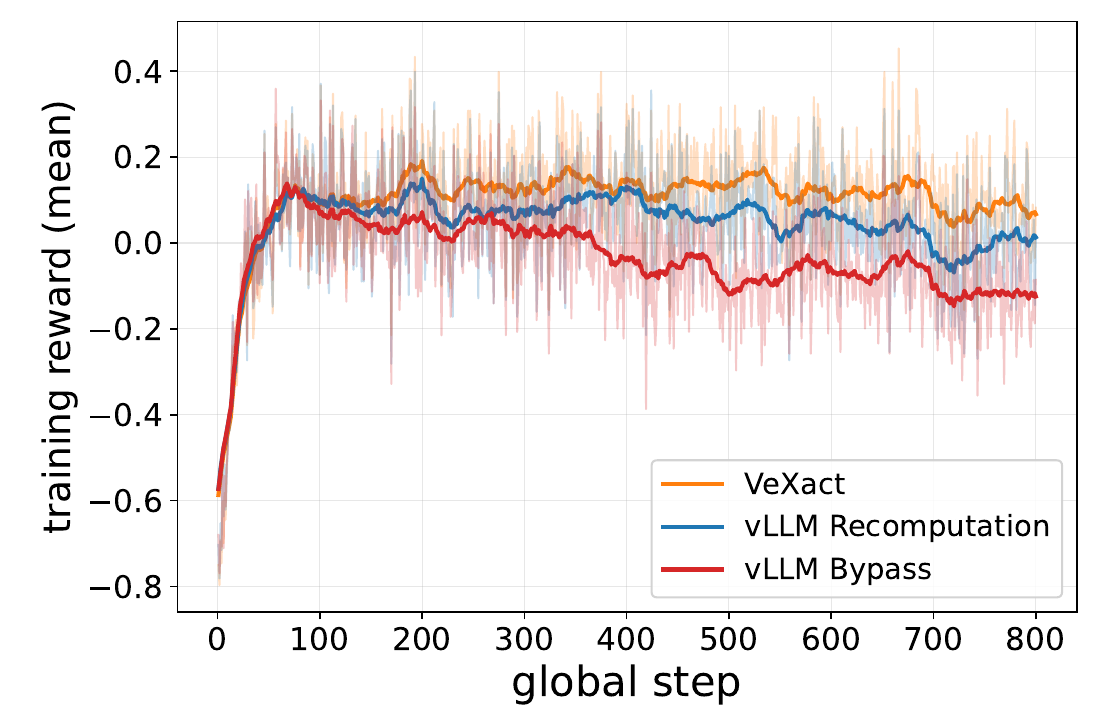}
    \caption{Training reward.}
    \end{subfigure}
    \begin{subfigure}[t]{0.34\textwidth}
    \centering
    \includegraphics[width=0.99\linewidth]{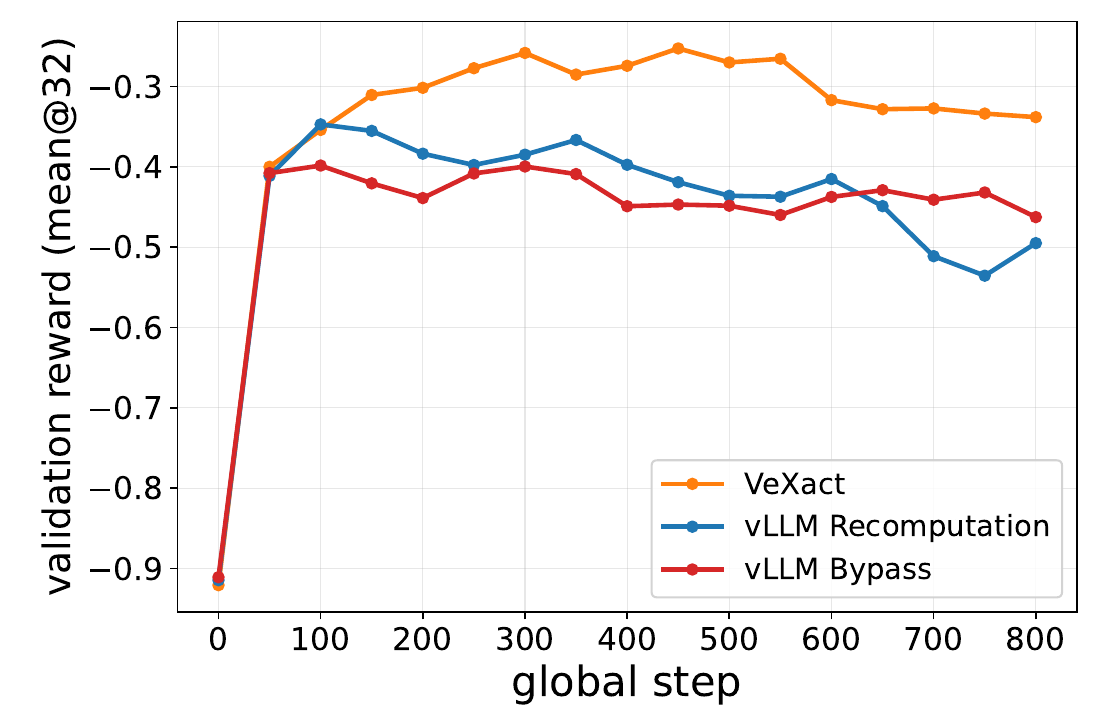}
    \caption{Validation reward.}
    \end{subfigure}
    \caption{Qwen3-1.7B GRPO experiments on the DAPO dataset, with \sysname{} and vLLM recomputation and bypass, where \sysname{} can maintain the training stability better. }
    \label{fig:dense_dapo}
\end{figure}

\begin{figure}[htb]
    \centering
    \begin{subfigure}[t]{0.34\textwidth}
    \centering
    \includegraphics[width=0.99\linewidth]{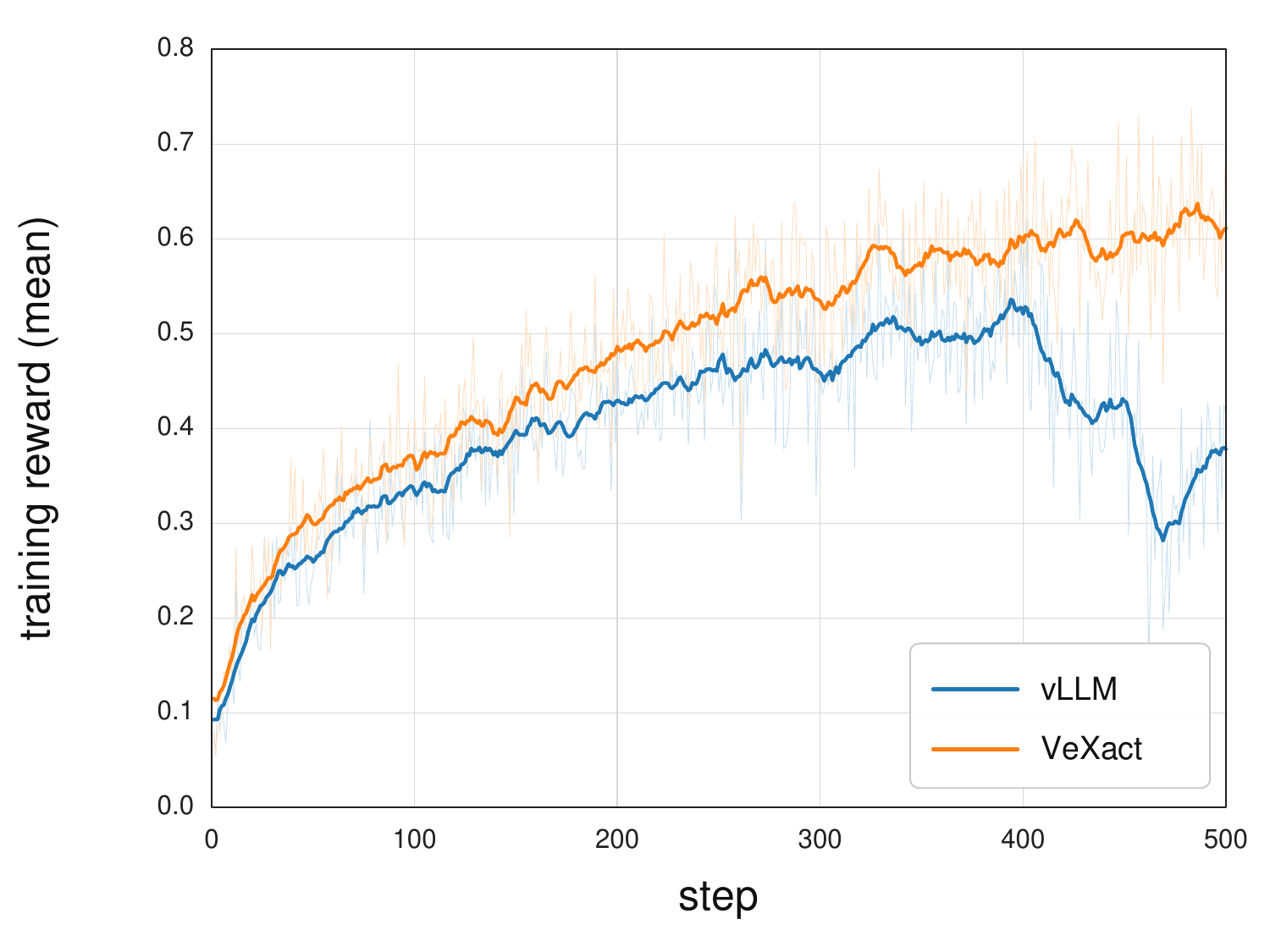}
    \caption{Comparison on Llama 3.2 model.}
    \end{subfigure}
    \begin{subfigure}[t]{0.34\textwidth}
    \centering
    \includegraphics[width=0.99\linewidth]{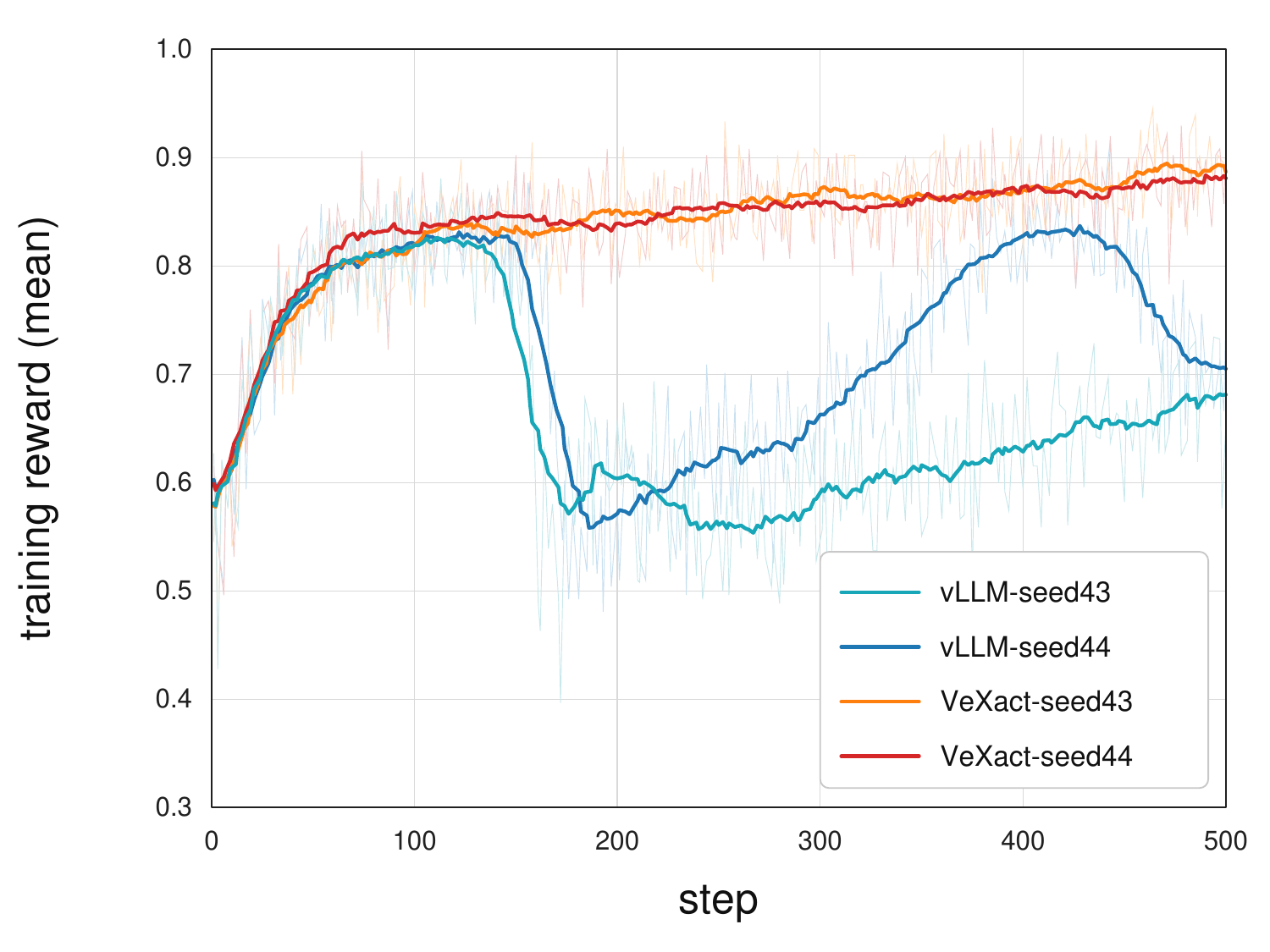}
    \caption{Comparison on different seeds.}
    \end{subfigure}
    \caption{Additional GRPO experiments on Llama 3.2 and across different random seeds, comparing \sysname{} with vLLM. All preceding experiments use random seed 42, we additionally report results with seeds 43 and 44 here. Across all seeds, the results support the same conclusion as in the paper: \sysname{} consistently maintains stable training.}
    \label{fig:llama_dapo}
\end{figure}

\subsection{Expanded Objectives for Post-hoc Patch Variants in Section~\ref{subsec:rollout_correction}}
\label{app:patch_objectives}

Let \(r^{\mathrm{rollout}}_{\mathrm{ppo},t}\) be the rollout-side PPO ratio
defined in Eq.~\ref{eq:ppo_ratio}, and let \(\mathcal{L}_{\mathrm{PPO}}\)
denote the token-level clipped surrogate. 

\paragraph{srs-k3-corr-ratio.}
Sequence-level rejection using the correction ratio
\(r_{\mathrm{corr}}\) as the filtering signal:
\begin{equation}
\mathcal{L}_{\mathrm{srs\text{-}k3\text{-}corr}}
=
\mathbf{1}
\left[
\sum_{t=1}^{T}K3(r_{\mathrm{corr},t})
\le \tau_{\mathrm{seq}}
\right]
\sum_{t=1}^{T}
\mathcal{L}_{\mathrm{PPO}}
\left(
r^{\mathrm{rollout}}_{\mathrm{ppo},t}, A_t
\right).
\label{eq:srs_corr_loss}
\end{equation}

\paragraph{srs-k3-ppo-ratio.}
Sequence-level rejection using the rollout-side PPO ratio
\(r^{\mathrm{rollout}}_{\mathrm{ppo}}\) as the filtering signal:
\begin{equation}
\mathcal{L}_{\mathrm{srs\text{-}k3\text{-}ppo}}
=
\mathbf{1}
\left[
\sum_{t=1}^{T}K3(r^{\mathrm{rollout}}_{\mathrm{ppo},t})
\le \tau_{\mathrm{seq}}
\right]
\sum_{t=1}^{T}
\mathcal{L}_{\mathrm{PPO}}
\left(
r^{\mathrm{rollout}}_{\mathrm{ppo},t}, A_t
\right).
\label{eq:srs_ppo_loss}
\end{equation}

\paragraph{tis-srs-k3-corr-ratio.}
Token-level truncation followed by sequence-level rejection, using the
correction ratio \(r_{\mathrm{corr}}\) as the filtering signal:
\begin{equation}
\mathcal{L}_{\mathrm{tis\text{-}srs\text{-}k3\text{-}corr}}
=
\mathbf{1}
\left[
\sum_{t=1}^{T}K3(r_{\mathrm{corr},t})
\le \tau_{\mathrm{seq}}
\right]
\sum_{t=1}^{T}
\min
\left(
r_{\mathrm{corr},t},
\tau_{\mathrm{tok}}
\right)
\mathcal{L}_{\mathrm{PPO}}
\left(
r^{\mathrm{train}}_{\mathrm{ppo},t}, A_t
\right).
\end{equation}

\paragraph{tis-srs-k1-corr-ratio.}
Token-level truncation followed by sequence-level rejection, using the
correction ratio \(r_{\mathrm{corr}}\) as the filtering signal:
\begin{equation}
\mathcal{L}_{\mathrm{tis\text{-}srs\text{-}k1\text{-}corr}}
=
\mathbf{1}
\left[
\sum_{t=1}^{T}K1(r_{\mathrm{corr},t})
\le \tau_{\mathrm{seq}}
\right]
\sum_{t=1}^{T}
\min
\left(
r_{\mathrm{corr},t},
\tau_{\mathrm{tok}}
\right)
\mathcal{L}_{\mathrm{PPO}}
\left(
r^{\mathrm{train}}_{\mathrm{ppo},t}, A_t
\right).
\end{equation}

\paragraph{Threshold choice.}
The effectiveness of post-hoc algorithmic compensation depends on the token-level truncation and sequence-level rejection thresholds. We follow settings adopted in prior work~\citep{qiu2026fp8rl,li2026mis}. Specifically, we use \(\tau_{\mathrm{tok}}=2\) for token-level truncated importance sampling and \(\tau_{\mathrm{seq}}=0.001\) for sequence-level mismatch rejection. Our claim is not that these thresholds are suboptimal. Rather, without a zero-mismatch reference, threshold selection based only on observed training performance does not reveal how closely a correction recovers mismatch-free behavior. \sysname{} provides such a reference and enables principled calibration of post-hoc corrections.


\end{document}